\documentclass[journal]{IEEEtran}
\usepackage{amsmath, amsfonts}
\allowdisplaybreaks

\usepackage{algorithmic}
\usepackage{algorithm}
\usepackage{array}
\usepackage[caption=false,font=normalsize,labelfont=sf,textfont=sf]{subfig}
\usepackage{textcomp}
\usepackage{stfloats}
\usepackage{url}
\usepackage{verbatim}
\usepackage{graphicx}
\usepackage{cite}
\usepackage{multirow, booktabs}
\usepackage{graphicx}
\usepackage{xcolor}

\usepackage[normalem]{ulem}
\useunder{\uline}{\ul}{}
\usepackage{amssymb}
\usepackage[hidelinks]{hyperref}
\usepackage{threeparttable}
\usepackage{makecell}
\usepackage{newtxtext,newtxmath}
\usepackage{siunitx}
\usepackage{todonotes}

\begin{document}
\title{Spiking Neural Networks for Temporal Processing: Status Quo and Future Prospects}

\author{ Chenxiang~Ma{*}, Xinyi Chen{*}, Yanchen Li{*}, Qu Yang, Yujie~Wu, Guoqi~Li,~\IEEEmembership{Member,~IEEE}, Gang~Pan,~\IEEEmembership{Senior Member,~IEEE}, Huajin~Tang,~\IEEEmembership{Senior~Member,~IEEE}, Kay~Chen~Tan,~\IEEEmembership{Fellow,~IEEE}, Jibin~Wu,~\IEEEmembership{Member,~IEEE}
\IEEEcompsocitemizethanks{
\IEEEcompsocthanksitem  *Chenxiang~Ma, Xinyi~Chen, and Yanchen~Li contributed equally to this article. Corresponding Author: Jibin~Wu (jibin.wu@polyu.edu.hk)
\IEEEcompsocthanksitem{Chenxiang~Ma, Xinyi~Chen, Yanchen~Li, Kay Chen Tan, and Jibin Wu are with the Department of Data Science and Artificial Intelligence, The Hong Kong Polytechnic University, Hong Kong SAR. Jibin Wu is also with the Department of Computing, The Hong Kong Polytechnic University, Hong Kong SAR}
\IEEEcompsocthanksitem{Qu Yang is with the Department of Electrical and Computer Engineering, National University of Singapore, Singapore 119077}
\IEEEcompsocthanksitem{Yujie Wu is with the Department of Computing, The Hong Kong Polytechnic University, Hong Kong SAR}
\IEEEcompsocthanksitem{Guoqi Li is with the Institute of Automation, Chinese Academy of Sciences, Beijing 100045, China}
\IEEEcompsocthanksitem{Gang Pan and Huajin Tang are with the State Key Laboratory of Brain-Machine Intelligence, College of Computer Science and Technology, MOE Frontier Science Center for Brain Science and Brain-Machine Integration, Zhejiang University, Hangzhou 310027, China}

}
}

\maketitle

\begin{abstract}
Temporal processing is fundamental for both biological and artificial intelligence systems, as it enables the comprehension of dynamic environments and facilitates timely responses. Spiking Neural Networks (SNNs) excel in handling such data with high efficiency, owing to their rich neuronal dynamics and sparse activity patterns. Given the recent surge in the development of SNNs, there is an urgent need for a comprehensive evaluation of their temporal processing capabilities. In this paper, we first conduct an in-depth assessment of commonly used neuromorphic benchmarks, revealing critical limitations in their ability to evaluate the temporal processing capabilities of SNNs. To bridge this gap, we further introduce a benchmark suite consisting of three temporal processing tasks characterized by rich temporal dynamics across multiple timescales. Utilizing this benchmark suite, we perform a thorough evaluation of recently introduced SNN approaches to elucidate the current status of SNNs in temporal processing. Our findings indicate significant advancements in recently developed spiking neuron models and neural architectures regarding their temporal processing capabilities, while also highlighting a performance gap in handling long-range dependencies when compared to state-of-the-art non-spiking models. Finally, we discuss the key challenges and outline potential avenues for future research.
\end{abstract}


\section{Introduction}
\IEEEPARstart{T}{emporal} processing is a fundamental capability that enables animals to perceive, plan, and act within dynamic environments. To effectively analyze and understand temporal data, a variety of models have been developed based on artificial neural networks (ANNs), such as Recurrent Neural Networks (RNNs)~\cite{RNN,LSTM}, Temporal Convolutional Networks (TCNs)~\cite{TCN}, Transformers~\cite{transformer}, and State Space Models~(SSMs)~\cite{SSM}. Despite their impressive performance in processing temporal data, these models often demand significant computational resources, which can limit their deployment on resource-constrained platforms~\cite{9043731}.

 
In contrast, Spiking Neural Networks (SNNs)~\cite{maass1997networks}, inspired by the computational principles of biological neural networks, offer an energy-efficient computational framework for temporal processing~\cite{10636118}. By employing a sparse spike-based representation, SNNs enable efficient event-driven computation, wherein spiking neurons activate solely in response to incoming spikes~\cite{roy2019towards,rapid2013}. This characteristic is particularly advantageous when implemented on neuromorphic chips~\cite{davies2018loihi,pei2019towards,ma2017darwin,Darwin3,yao2024nc}, where SNNs have demonstrated significantly enhanced energy efficiency compared to traditional ANNs~\cite{10.1162/neco_a_01245}. Beyond their energy efficiency, spiking neurons inherently function as stateful models, characterized by rich neuronal dynamics that arise from complex morphology, variations in ionic conductance, and the distribution of synaptic inputs~\cite{herz2006modeling}. These features endow SNNs with significant potential for representing and processing temporal data~\cite{wu2020deep,wu2018spiking,sequence2016}.

\begin{figure*}[!t]
\centering\includegraphics[width=0.95\linewidth]{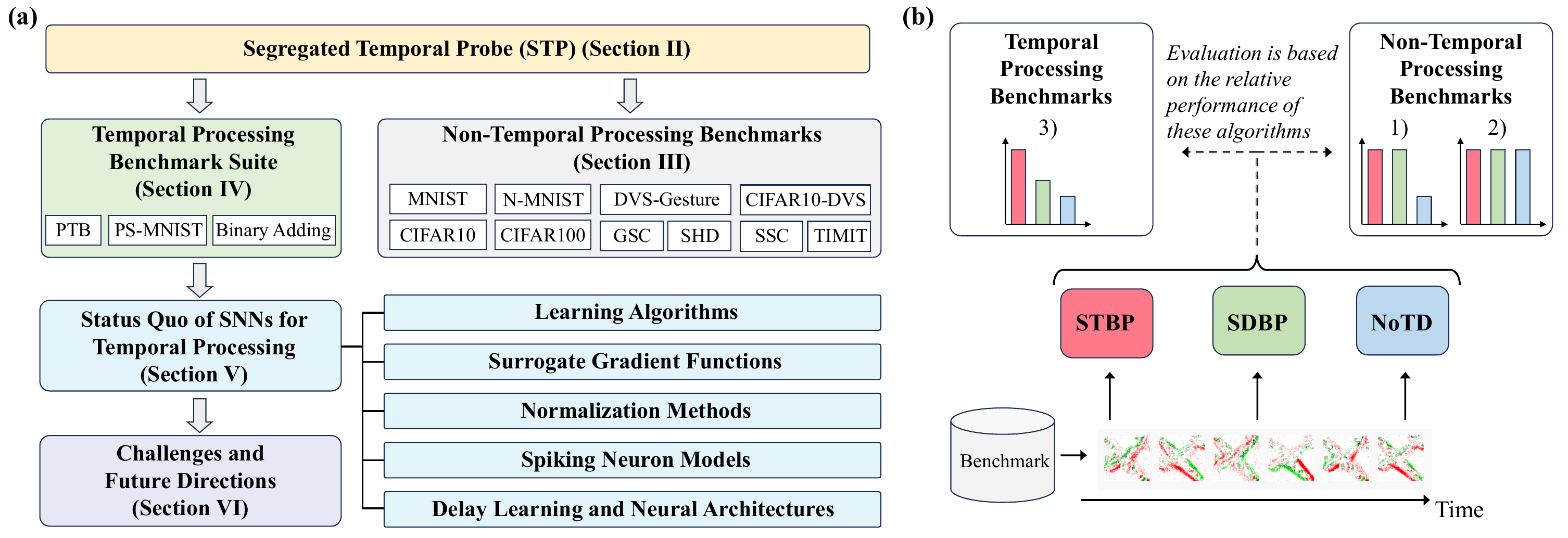}
\caption{(a) Overview of the paper organization. In Section~\ref{sec:STP}, we propose the Segregated Temporal Probe (STP) analytical tool for assessing the effectiveness of neuromorphic benchmarks in evaluating the temporal processing capabilities of SNN. In Section~\ref{sec:eval_bench}, using the STP, we discover that commonly used neuromorphic benchmarks are ineffective for assessing temporal processing performance. Then, we introduce a suite of temporal processing benchmarks in Section~\ref{sec:benchmark}. Based on this suite, we conduct a comprehensive benchmarking study to reveal the current status of SNNs for temporal processing in Section~\ref{sec:benchmarking}. Finally, we discuss key challenges and outline future directions in Section~\ref{sec:challenges}. (b) Illustration of the proposed STP. STP incorporates three algorithms (STBP, SDBP, and NoTD) that systematically disrupts the temporal processing pathways within an SNN to elucidate their significance. }
\label{Fig:overview}
\end{figure*}

Recently, numerous approaches have been developed to enhance training efficiency~\cite{ptl,NEURIPS2022_523caec7,10254579,Xu_2023_CVPR,deng2022temporal,imloss,ijcai2023p335,zhu2024online,jiang2024ndot,shen2024rethinking} and representation power of SNNs~\cite{hybridcoding,attentionsnns,NEURIPS2023_b8734840,9328792,ternarySpike2024,xing2024spikelm,hu2024toward}. Despite these advancements, a clear consensus on how to evaluate these approaches in the context of temporal processing remains elusive. Additionally, the lack of standardized training and evaluation configurations across these studies complicates fair comparisons.
These challenges collectively impede the progress of the field and limit the real-world applicability of SNNs.
In this paper, we seek to address these issues by establishing a comprehensive evaluation benchmark specifically focused on temporal processing capabilities. Furthermore, we will assess existing approaches to elucidate the current state of the field and identify potential future research directions. An overview of the paper's structure is presented in Fig.~\ref{Fig:overview}.

Benchmarks are essential in Artificial Intelligence research, as they provide standardized datasets and evaluation metrics that facilitate consistent performance comparison, track progress, and promote reproducibility. Currently, the benchmarks commonly employed for SNN evaluation can be categorized into three groups. The first category comprises static image recognition datasets~\cite{lecun1998gradient, krizhevsky2009learning}, which require the conversion of static images into sequences, typically by replicating images along the time axis~\cite{eshraghian2023training}. The second category encompasses event-based vision datasets generated using neuromorphic sensors, such as Dynamic Vision Sensor (DVS) cameras. Some of these datasets are created by imposing artificial saccadic motion on static images~\cite{orchard2015converting, li2017cifar10}, while others directly capture real-world moving objects using DVS cameras~\cite{amir2017low,zhou2024enhancing}. The third category consists of audio classification datasets, which are either created using the original audio signals~\cite{warden2018speech,TIMIT} or converted into spike-based representations using biologically inspired encoding methods~\cite{pan2020efficient,cramer2020heidelberg}. While these benchmarks have significantly advanced neuromorphic computing research over the past decade~\cite{10636118,eshraghian2023training}, their effectiveness in evaluating the temporal processing capability remains unclear, potentially leading to inaccurate assessments and misleading conclusions. 

To address this issue, we propose an analytical tool called the Segregated Temporal Probe (STP),  designed to evaluate the effectiveness of a benchmark dataset in assessing the temporal processing capabilities of SNNs. Specifically, STP incorporates three learning algorithms: Spatio-Temporal Backpropagation (STBP)~\cite{wu2018spatio}, Spatial Domain Backpropagation (SDBP), and No Temporal Domain~(NoTD). These algorithms systematically disrupt the temporal processing pathways within an SNN to elucidate their significance. In particular, STBP preserves the full temporal processing pathways of the SNN during both forward and backward passes, whereas SDBP stops gradient propagation along the temporal domain during the backward pass. In contrast, NoTD stops the propagation of information along the temporal domain during both passes, effectively treating each time step independently.

By applying STP to widely adopted neuromorphic benchmark datasets, we find that NoTD achieves performance comparable to STBP on static image recognition~\cite{lecun1998gradient, krizhevsky2009learning} and event-based vision datasets~\cite{orchard2015converting, li2017cifar10,amir2017low}. This finding suggests that these datasets can be effectively processed without relying on temporal processing capabilities of SNN models. For audio classification datasets~\cite{warden2018speech,cramer2020heidelberg,TIMIT}, both SDBP and STBP obtain similar performance, suggesting that temporal credit assignment during the backward pass is not necessary for these datasets. Consequently, these existing benchmark datasets do not adequately evaluate the temporal processing capability of SNNs.

To bridge this gap and elucidate the status quo of the temporal processing capabilities of existing SNN approaches, we introduce a benchmark suite encompassing three temporal processing tasks characterized by rich temporal dynamics. Subsequently, we conduct a comprehensive evaluation of over thirty SNN approaches using this benchmark suite. Our benchmarking study reveals three significant findings that have not been previously reported: (1) Online learning algorithms~\cite{xiao2022online,meng2023towards}, which claim performance comparable to or even superior to STBP, often yield less competitive results on temporal processing tasks. This indicates that the omitted temporal gradients are crucial for learning temporal dependencies within such data. (2) Surrogate gradient functions~\cite{neftci2019surrogate} with smoother curves and reduced gradient mismatching, such as Triangle~\cite{deng2022temporal} and Sigmoid~\cite{adLIF}, prove to be more effective for temporal processing tasks. (3) Recent advancements in spiking neuron models that incorporate strategies such as additional memory states~\cite{TCLIF,LMH}, heterogeneous neuron parameters~\cite{PLIF,GLIF}, and enriched recurrent neuron dynamics~\cite{PMSN, ltc,dhsnn} demonstrate significant improvements over the simplified Leaky Integrated-and-Fire~(LIF) model~\cite{lif} in temporal processing. However, despite their considerable energy efficiency, these state-of-the-art (SOTA) SNN models still lag behind ANN models~\cite{LSTM,SSM} in their ability to model long-range temporal dependencies.

Our major contributions in this work are summarized as follows: 
\begin{itemize}
\item We propose an analytical tool, STP, for evaluating the effectiveness of neuromorphic benchmarks in assessing temporal processing capabilities. This tool facilitates the development of neuromorphic benchmarks specifically tailored for temporal processing.
\item We identify critical limitations in existing neuromorphic benchmarks regarding their evaluation of temporal processing capabilities and propose a new benchmark suite explicitly designed for this purpose. 
\item We conduct a comprehensive benchmarking study of over thirty SNN approaches to elucidate the current state of the field.
\item We develop an open-sourced library\footnote{Code is publicly available at \url{https://github.com/liyc5929/neuroseqbench}.} for neuromorphic temporal processing, which enables consistent performance comparisons across different approaches and facilitates the tracking of advancements in the field.
\end{itemize}

The remainder of this paper is organized as follows. Section~\ref{sec:STP} introduces the proposed STP tool, which is utilized to evaluate existing neuromorphic benchmarks for temporal processing in Section~\ref{sec:eval_bench}.  Section~\ref{sec:benchmark} presents a novel neuromorphic benchmark suite specifically designed for temporal processing. Subsequently, we conduct a comprehensive evaluation of over thirty SNN approaches in Section~\ref{sec:benchmarking} to elucidate the current state of the field. Section~\ref{sec:challenges} discusses key challenges and outlines potential future research directions. Finally, we conclude the paper in Section~\ref{sec:conclusion}.

\begin{figure}[!t]
\centering\includegraphics[width=0.9\linewidth]{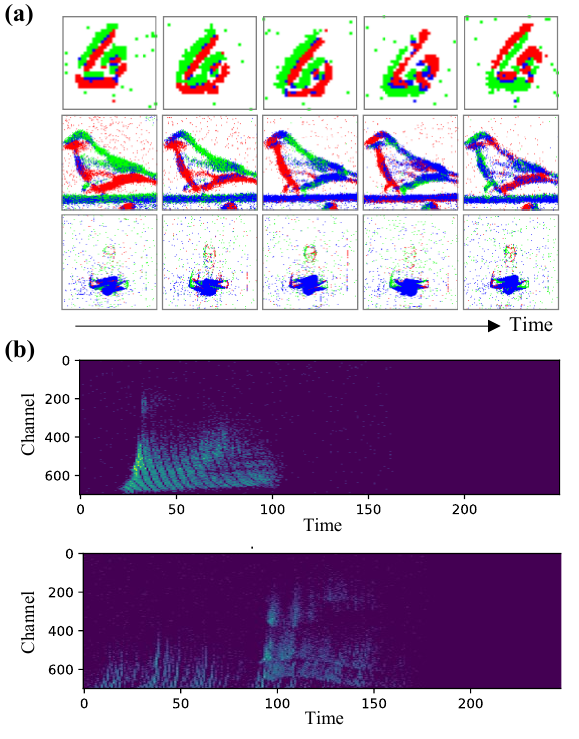}
\caption{Visualization of samples in neuromorphic benchmarks. (a) Samples from event-based vision datasets: N-MNIST, CIFAR10-DVS, and DvsGesture (from top to bottom). (b) Samples from neuromorphic audio datasets: SHD (top) and SSC (bottom).}
\label{Fig:illustration_samples}
\end{figure}

\begin{figure}[!t]
\centering\includegraphics[width=0.96\linewidth, trim = 0 0 0 0, clip]{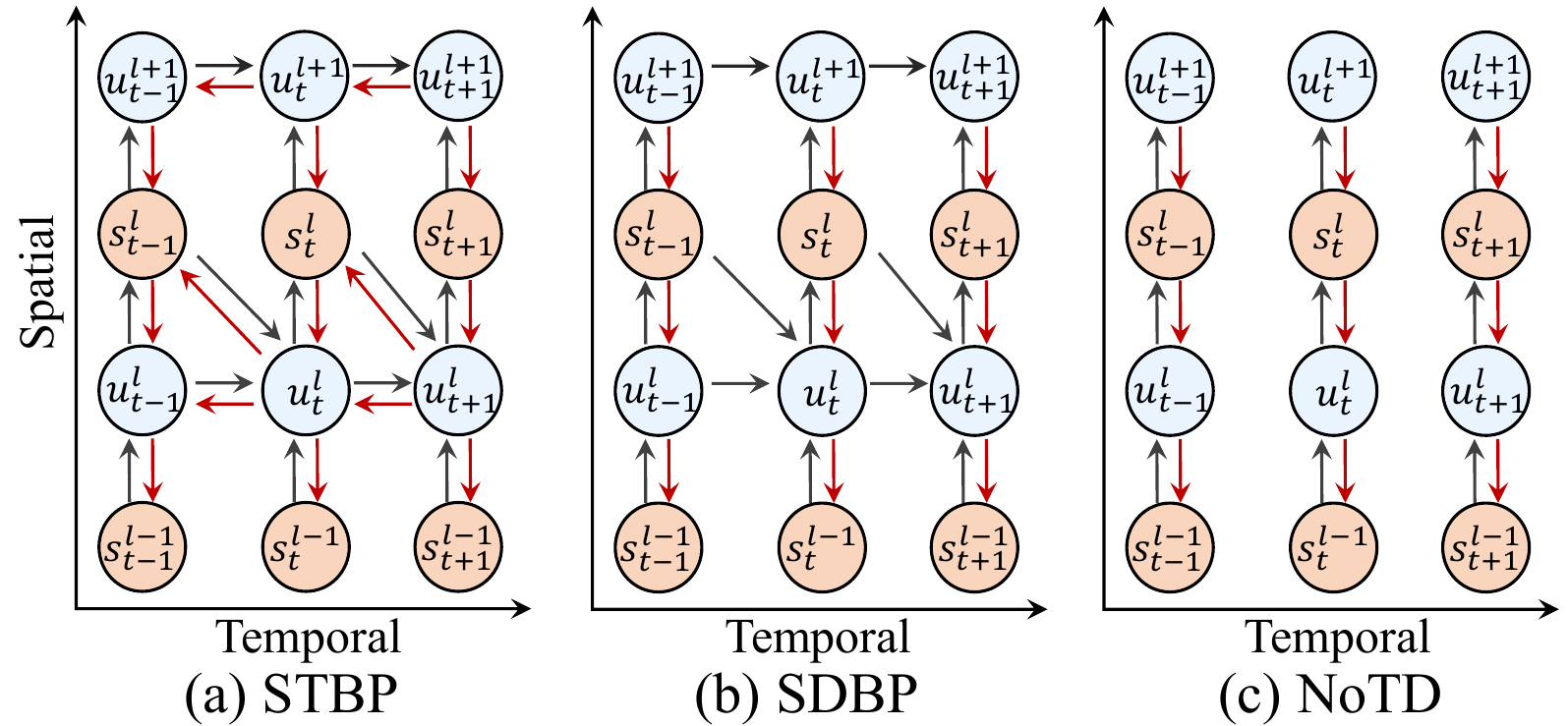}
\caption{Comparison of the computational graphs for three algorithms utilized in the STP. Forward and backward passes are denoted by black and red arrows, respectively.} 
\label{fig:stp}
\end{figure}

\section{Segregated Temporal Probe (STP)}
\label{sec:STP}

While existing neuromorphic benchmarks have significantly advanced the field of neuromorphic computing~\cite{10636118,eshraghian2023training}, it remains unclear whether these benchmarks effectively evaluate critical temporal information. 
This uncertainty is particularly evident in benchmarks adapted from static datasets~\cite{orchard2015converting, li2017cifar10}, where objects in individual frames often provide sufficient information for classification, as illustrated in Fig.~\ref{Fig:illustration_samples}(a). Consequently, we are motivated to explicitly analyze whether temporal processing capability is genuinely critical for achieving high performance on these benchmarks. To this end, we introduce the STP, which disrupts the temporal processing pathways within an SNN to elucidate their significance.

As illustrated in Fig.~\ref{Fig:overview}(b), the STP tool comprises three learning algorithms: STBP~\cite{wu2018spatio}, SDBP, and NoTD. STBP serves as the baseline, retaining the entire temporal processing pathways, including both the forward pass of activation values and the backward pass of error gradients across time. In contrast, SDBP is designed to disrupt temporal processing during the backward pass while preserving it in the forward pass. NoTD, on the other hand, eliminates temporal processing entirely by processing each frame independently at each time step. In the following, we will use the LIF neuron model as an example to illustrate these algorithms. Fig.~\ref{fig:stp} provides a visualization that highlights the differences in the 
forward and backward passes of each algorithm. 

\paragraph{LIF Neuron Model} The LIF neuron model~\cite{lif} is one of the most widely used spiking neuron models due to its simplicity and analytical tractability~\cite{10636118}. As described in Eqs.~\eqref{eq:mem_update}--\eqref{eq:firing}, LIF neurons capture the dynamics of membrane potential, which continuously integrates input spikes from preceding neurons. When the membrane potential exceeds a specified firing threshold, a spike is generated, followed by a reset of the membrane potential to its resting state. 
\begin{align}
\boldsymbol{u}^{l}[t] &= \underbrace{\lambda \cdot \boldsymbol{u}^{l}[t - 1] \cdot (1 - \boldsymbol{s}^{l}[t-1])}_{\text{temporal processing}} +\, \boldsymbol{W}^{l} \cdot \boldsymbol{s}^{l-1}[t], \label{eq:mem_update}\\
\boldsymbol{s}^{l}[t]  &= \Theta (\boldsymbol{u}^{l}[t] - V_{\text{th}}), \\[0.7ex]
\Theta(x)&=\begin{cases}1, & x \geq 0, \\[0.5ex]
                        0, & \text{otherwise,}\end{cases}
\label{eq:firing}
\end{align}
where $\boldsymbol{u}^{l}[t]$ represents the membrane potential in layer $l$ at time step~$t$, $\lambda$ denotes a decay factor that determines the rate at which the membrane potential decays over time, and $\boldsymbol{s}^{l-1}[t]$ represents input spikes from the previous layer. $\boldsymbol{W}^{l}$ is the weight matrix, $\Theta(x)$ is the Heaviside step function, and $V_{\text{th}}$ denotes the firing threshold. The resting potential is typically set to zero in practice.

\paragraph{STBP} STBP is a gradient-based learning algorithm specifically designed for SNNs~\cite{wu2018spatio}. After input spikes are propagated to the output layer $L$, a loss $\mathcal{L}$ is calculated by comparing the predictions with the target values. The backward pass then initiates, calculating the gradient of the loss with respect to each model parameter. By employing the chain rule, the gradient of the loss with respect to the weights of layer $l$ can be computed as follows:
\begin{align}
\!\frac{\partial \mathcal{L}}{\partial \boldsymbol{W}^{l}} &= \sum_{t=1}^{T}\frac{\partial \mathcal{L}}{\partial \boldsymbol{u}^{l}[t]} \cdot \frac{\partial \boldsymbol{u}^{l}[t]}{\partial \boldsymbol{W}^{l}} = \sum_{t=1}^{T}{\boldsymbol{\delta}^{l}[t]}^{\!\top} \cdot {\boldsymbol{s}^{l-1}[t]}^{\!\top}, \label{eq:weight}
\\[1.5ex]
\boldsymbol{\delta}^{l}[t]\!&=\!\begin{cases} 
\frac{\partial \mathcal{L}}{\partial \boldsymbol{u}^L[T]},  &l\!=\!L~\text{and}~t\!=\!T, \\[1.5ex] 
\boldsymbol{\delta}^{L}[t\!+\!1] \!\cdot\! \frac{\partial \boldsymbol{u}^{L}[t+1]}{\partial \boldsymbol{u}^{L}[t]}+\frac{\partial \mathcal{L}}{\partial \boldsymbol{u}^L[t]},  &l\!=\!L~\text{and}~t\!<\!T, \\[1.5ex] 
\boldsymbol{\delta}^{l+1}[T] \!\cdot\! \frac{\partial \boldsymbol{u}^{l+1}[T]}{\partial \boldsymbol{s}^{l}[T]} \!\cdot\! \frac{\partial \boldsymbol{s}^{l}[T]}{\partial \boldsymbol{u}^{l}[T]},  &l\!<\!L~\text{and}~t\!=\!T, \\[1.5ex] 
\boldsymbol{\delta}^{l}\![t\!+\!1]\frac{\partial \boldsymbol{u}^{l}\![t+1]}{\partial \boldsymbol{u}^{l}\![t]} + \boldsymbol{\delta}^{l+1}\![t] \frac{\partial \boldsymbol{u}^{l+1}\![t]}{\partial \boldsymbol{u}^{l}\![t]},\!\!\!&\!\text{otherwise,}\! \end{cases}
\label{eq:grad_m}
\end{align}
where $\boldsymbol{\delta}^{l}[t]\triangleq\frac{\partial \mathcal{L}}{\partial \boldsymbol{u}^{l}[t]}$, represents the error back-propagated from the last layer and the last time step. $T$ denotes the total number of time steps. The Heaviside step function is non-differentiable since its gradient, i.e., $\frac{\partial \boldsymbol{s}_{i}^{l}[t]}{\partial \boldsymbol{u}_{i}^{l}[t]}$, is zero except at the point where $\boldsymbol{u}_{i}^{l}[t]=V_{\text{th}}$, where it becomes infinite. To address this issue, a continuous surrogate gradient function~\cite{neftci2019surrogate} is often adopted to replace the non-differentiable Heaviside step function during the backward pass, as denoted by $\frac{\partial \boldsymbol{s}_i^{l}[t]}{\partial \boldsymbol{u}_i^{l}[t]} \approx \mathbb{H}\!\left(\boldsymbol{u}_i^{l}[t]\right)$.

\paragraph{SDBP} To illustrate SDBP, we begin by explicitly detailing the gradients associated with the temporal processing function. This is achieved by rewriting the error term $\boldsymbol{\delta}^{l}[t]$ under the general condition~($l<L~\text{and}~t<T$):
\begin{align}
\boldsymbol{\delta}^{l}[t] &= \boldsymbol{\delta}^{l+1}[t] \cdot \frac{\partial \boldsymbol{u}^{l+1}[t]}{\partial \boldsymbol{s}^{l}[t]} \cdot \frac{\partial \boldsymbol{s}^{l}[t]}{\partial \boldsymbol{u}^{l}[t]}\nonumber\\
&\phantom{=} + \underbrace{\sum_{t^\prime=t+1}^T
\boldsymbol{\delta}^{l+1}[t^{\prime}] \cdot \frac{\partial \boldsymbol{u}{^{l+1}}[t^\prime]}{\partial \boldsymbol{u}{^{l}}[t^\prime]} \cdot \frac{\partial \boldsymbol{u}{^{l}}[t^\prime]}{\partial \boldsymbol{u}{^{l}}[t]}}_{\text{temporal processing gradients}},
\label{eq:td_grad_m}
\end{align}
where the gradient at the current time step $t$ is influenced by a sum of the gradients from all subsequent time steps. This stems from the temporal processing function in Eq.~\eqref{eq:mem_update}. More concretely, the membrane potential at the current time step $t$ implicitly contributes to all membrane potentials at the subsequent time steps due to the decay and resetting processes over time. Accordingly, the errors at all subsequent steps need to be included in the computation of the error in the time $t$. 

SDBP keeps the same temporal processing function in the forward pass as STBP but omits it in the backward pass. Consequently, error signals generated at a given time step cannot be propagated back to previous time steps, limiting the use of error information to refine earlier predictions.  Formally, let  $\boldsymbol{\epsilon}^{l}[t]$ represent $\frac{\partial \mathcal{L}}{\partial \boldsymbol{u}^{l}[t]}$ in SDBP, the weight gradient at the layer $l$ is given by
\begin{align}
\nabla_{\boldsymbol{W}^{l}}\mathcal{L} &= \sum_{t=1}^T{\boldsymbol{\epsilon}^l[t]}^{\!\top} \cdot \frac{\partial \boldsymbol{u}^{l}[t]}{\partial\boldsymbol{W}^{l}} = \sum_{t=1}^T{\boldsymbol{\epsilon}^l[t]}^{\!\top} \cdot {\boldsymbol{s}^{l-1}[t]}^{\!\top},
\label{eq:tlg} 
\\[1.5ex]
\boldsymbol{\epsilon}^l[t] &= \begin{cases} 
\frac{\partial \mathcal{L}}{\partial \boldsymbol{u}^L[t]},  &l\!=\!L,\\[1.5ex] 
\boldsymbol{\epsilon}^{l+1}[t] \cdot \frac{\partial \boldsymbol{u}^{l+1}[t]}{\partial \boldsymbol{u}^{l}[t]},  &l\!<\!L.
\end{cases}
\label{eq:tlg2}
\end{align}

\begin{table}[!t]
\centering
\caption{Evaluation results of widely-used neuromorphic benchmarks. ``Acc." stands for ``accuracy".}
\setlength{\tabcolsep}{1.5mm}
\begin{tabular}{l|c|ccc}
\hline
\textbf{Dataset} & \textbf{Time Step} (\(T\))   & \textbf{Method} & \textbf{Acc.} & \(\Delta\)\textbf{Acc.} \\ \hline
\multirow{3}{*}{MNIST~\cite{lecun1998gradient}} & \multirow{3}{*}{10}      & STBP   & 99.40    &  -      \\
                             && SDBP   & 99.27    & -0.13  \\
                             && NoTD   & 99.18    & -0.22  \\ \hline
\multirow{3}{*}{CIFAR10~\cite{krizhevsky2009learning}} & \multirow{3}{*}{4}     & STBP   & 94.86    & -       \\
                             && SDBP   & 94.74    & -0.12  \\
                             && NoTD   & 93.46    & -1.40  \\ \hline
\multirow{3}{*}{CIFAR100~\cite{krizhevsky2009learning}} & \multirow{3}{*}{4}     & STBP   & 74.57    & -       \\
                             && SDBP   & 74.35    & -0.22  \\
                             && NoTD   & 73.28    & -1.29  \\ \hline
\multirow{3}{*}{N-MNIST~\cite{orchard2015converting}} & \multirow{3}{*}{300}   & STBP   &  99.49   &  -      \\
                            & & SDBP   &  99.48   &  -0.01 \\
                            & & NoTD   &  99.09   &  -0.40 \\ \hline
\multirow{3}{*}{CIFAR10-DVS~\cite{li2017cifar10}}& \multirow{3}{*}{10}& STBP    &  78.50   &  -      \\
                             && SDBP   &  79.00   &  +0.50 \\
                             && NoTD   &  80.00   &  +1.50 \\ \hline
\multirow{3}{*}{DvsGesture~\cite{amir2017low}} &\multirow{3}{*}{20} & STBP   &  95.14   & -       \\
                             && SDBP   &  95.83   & +0.69  \\
                             && NoTD   &  94.44   & -0.70  \\ \hline
\multirow{3}{*}{GSC~\cite{warden2018speech}}   &\multirow{3}{*}{101}      & STBP   &  92.91   & -       \\
                             && SDBP   &  89.00   & -3.91  \\
                             && NoTD   &  77.53   & -15.38  \\ \hline
\multirow{3}{*}{SHD~\cite{cramer2020heidelberg}}  &\multirow{3}{*}{250}       & STBP   &  86.48   & -       \\
                             && SDBP   &  85.07   & -1.41  \\
                             && NoTD   &  68.51   & -17.97  \\ \hline
\multirow{3}{*}{SSC~\cite{cramer2020heidelberg}} &\multirow{3}{*}{250}        & STBP   &  67.13   & -       \\
                             && SDBP   &  66.03   & -1.10   \\
                             && NoTD   &  44.97   & -22.16   \\ \hline
\multirow{3}{*}{TIMIT~\cite{TIMIT}} &\multirow{3}{*}{100}         & STBP   &  57.07   & -       \\
                             && SDBP   &  53.24   & -3.83   \\
                             && NoTD   &  49.01   & -8.06   \\ \hline
\end{tabular}%
\label{tab:res_exis_bench}
\end{table}

\paragraph{NoTD} NoTD is designed to eliminate temporal processing functions in both forward and backward passes, allowing each time step in the input sequence to be processed independently without interaction with other time steps. Consequently, NoTD cannot learn temporal relationships within sequences or integrate information over time for decision-making. Formally, the forward pass of layer $l$ in NoTD is given by
\begin{align}
\boldsymbol{u}^{l}[t] &= \boldsymbol{W}^{l} \cdot \boldsymbol{s}^{l-1}[t], \label{eq:notd_forward}\\[1.5ex]
\boldsymbol{s}^{l}[t]  &= \Theta (\boldsymbol{u}^{l}[t] - V_{\text{th}}).
\end{align}


Compared to Eq.~\eqref{eq:mem_update}, Eq.~\eqref{eq:notd_forward} removes the temporal processing function, i.e., the membrane potential update process. 
In the backward pass, the weight gradient in NoTD is the same as Eqs.~\eqref{eq:tlg} and \eqref{eq:tlg2} in SDBP. 

\begin{figure*}[!t]
\centering\includegraphics[width=0.9\linewidth]{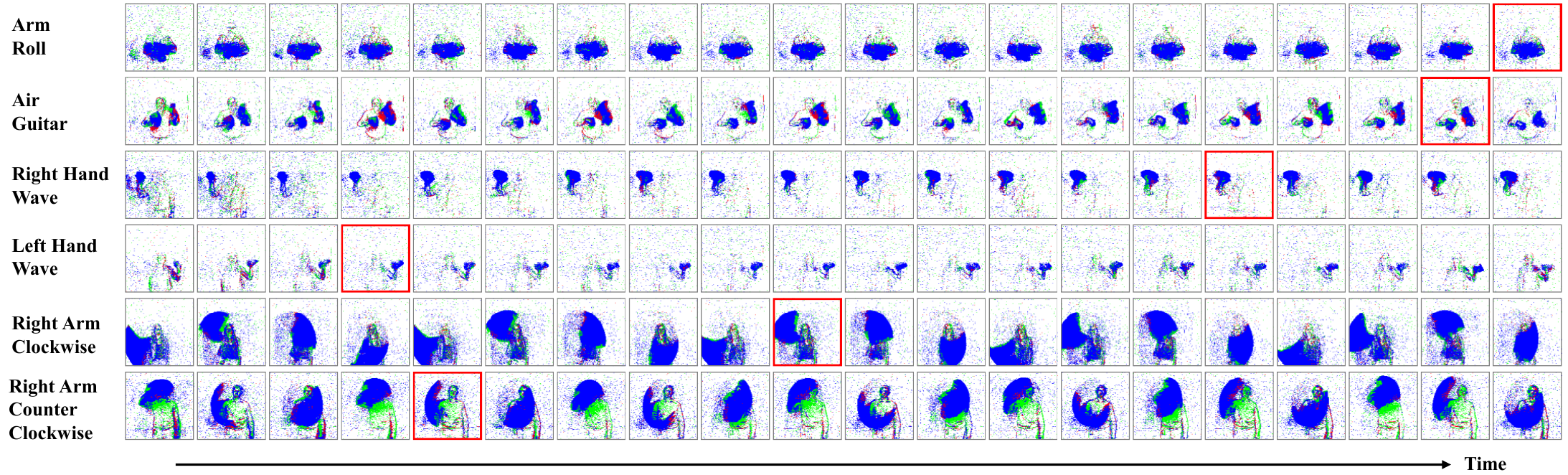}
\caption{Qualitative results of samples from the DvsGesture dataset, along with the confident frame (highlighted with red boxes) selected by all the algorithms in STP. Labels are provided in the leftmost column. }
\label{Fig:dvsgesture_vis}
\end{figure*}

\paragraph{Evaluation Criteria}
To determine whether a dataset can effectively evaluate the temporal processing capabilities of SNNs, we train SNNs using the three algorithms on the dataset. The effectiveness of the dataset will be evaluated based on the relative performance of these algorithms: 
\begin{enumerate}
  \item If the performance of SDBP is comparable to or exceeds that of STBP, this suggests that temporal credit assignment during the backward pass may be unnecessary, indicating that the dataset is not suitable for evaluating temporal processing capabilities.
  \item If the performance of NoTD is comparable to or better than that of SDBP, it implies that the dataset can be effectively addressed without incorporating temporal interactions within the model, thus rendering it an inadequate benchmark.
  \item Conversely, if SDBP surpasses NoTD and STBP outperforms SDBP, this indicates that information integration across the time is essential, and that temporal credit assignment effectively facilitate this integration. Therefore, the benchmark is suitable for assessing the temporal processing capabilities of SNNs.
\end{enumerate}

\section{Evaluation of Neuromorphic Benchmarks Using STP}
\label{sec:eval_bench}
We conduct a comprehensive analysis of the existing neuromorphic benchmarks using the STP to reveal their critical limitations. 
For each benchmark, we adhere to standard protocols from prior studies~\cite{wu2018spatio,xiao2022online,meng2023towards,duan2022temporal,ASGL,TCLIF}, which include dataset processing, data augmentation, network design, and training configurations. Detailed descriptions of these setups are provided in Appendix~\ref{app:imp_det_exsiting_bench}.

Our analysis begins with static image recognition benchmarks, including MNIST~\cite{lecun1998gradient}, CIFAR10~\cite{krizhevsky2009learning}, and CIFAR100~\cite{krizhevsky2009learning}. These datasets do not contain temporal information, as each input sequence is generated by repeating a static image along the time dimension. This is reaffirmed by our experimental results. As shown in Table~\ref{tab:res_exis_bench}, both NoTD and SDBP achieve accuracy comparable to STBP. This suggests that the capability to model temporal relationships is not necessary for high performance on these datasets.

Next, we examine event-based vision datasets. Results are presented in Table~\ref{tab:res_exis_bench}. For synthetic event-based datasets like N-MNIST~\cite{orchard2015converting} and CIFAR10-DVS~\cite{li2017cifar10}, NoTD performs similarly to or even outperforms both SDBP and STBP. This suggests that these datasets can be effectively addressed at the frame level without the need for temporal modeling. Interestingly, the DvsGesture dataset~\cite{amir2017low}, which captures human gestures in real time, is also effectively addressed by NoTD. This implies that the DvsGesture dataset does not require the model to have temporal processing capabilities for accurate recognition.

To understand these results, we conduct a qualitative analysis by visualizing samples from the DvsGesture dataset in Fig.~\ref{Fig:dvsgesture_vis}, along with the confident frame selected by each algorithm in STP. The confident frame is defined as the frame where the output layer has the strongest response. We observe that most samples have minimal changes over time, which allows accurate classification based on a single frame. The three algorithms often select the same confident frame, reinforcing that a single frame is sufficient for correct recognition without the need for temporal modeling. 

In cases where the algorithms choose different frames, as shown in Appendix Fig.~\ref{Fig:app_dvs_vis_diff_t}, the selected frames often have similar spatial features despite at different time steps. This demonstrates that these samples contain several informative frames adequate for pattern recognition. Additionally, certain classes, such as right arm clockwise and counterclockwise movements, seemingly necessitate temporal modeling due to their directional differences. However, both NoTD and SDBP successfully recognize these classes and select confident frames that align with STBP. This effectiveness can be attributed to the presence of distinguishable spatial features within these classes, which reduces the necessity for temporal modeling. Furthermore, we visualize samples correctly classified by STBP but not by NoTD in Appendix Fig.~\ref{Fig:app_dvs_vis_stbp_correct_notd_wrong}. We observe that the errors made by NoTD are primarily associated with spatial features rather than temporal cues. Similarly, as illustrated in Appendix Fig.~\ref{Fig:app_dvs_vis_stbpwrong}, the misclassifications by STBP are also linked to spatial features present within individual frames, rather than to temporal relationships across multiple frames. This analysis supports the conclusion that the primary challenge for achieving high-accuracy classification on the DvsGesture dataset resides in spatial pattern recognition rather than in temporal modeling.

Finally, we evaluate audio classification benchmarks, including GSC~\cite{warden2018speech}, SHD~\cite{cramer2020heidelberg}, SSC~\cite{cramer2020heidelberg}, and TIMIT~\cite{TIMIT}. Fig.~\ref{Fig:illustration_samples}(b) presents a visualization of randomly selected audio samples. By applying STP to these benchmarks, we find that a significant subset of samples can be accurately classified using only frame-level processing capabilities. This is evidenced by the moderate accuracy achieved by NoTD on these benchmarks. Furthermore, SDBP demonstrates substantially higher accuracy than NoTD, approaching the performance of STBP. This indicates that temporal interactions during the forward pass are critical for these tasks, while temporal credit assignment during the backward pass contributes only marginally to performance gains. Consequently, these benchmarks are not effective for evaluating temporal processing capabilities.

\begin{figure}[!t]
    \centering    \includegraphics[width=0.95\linewidth]{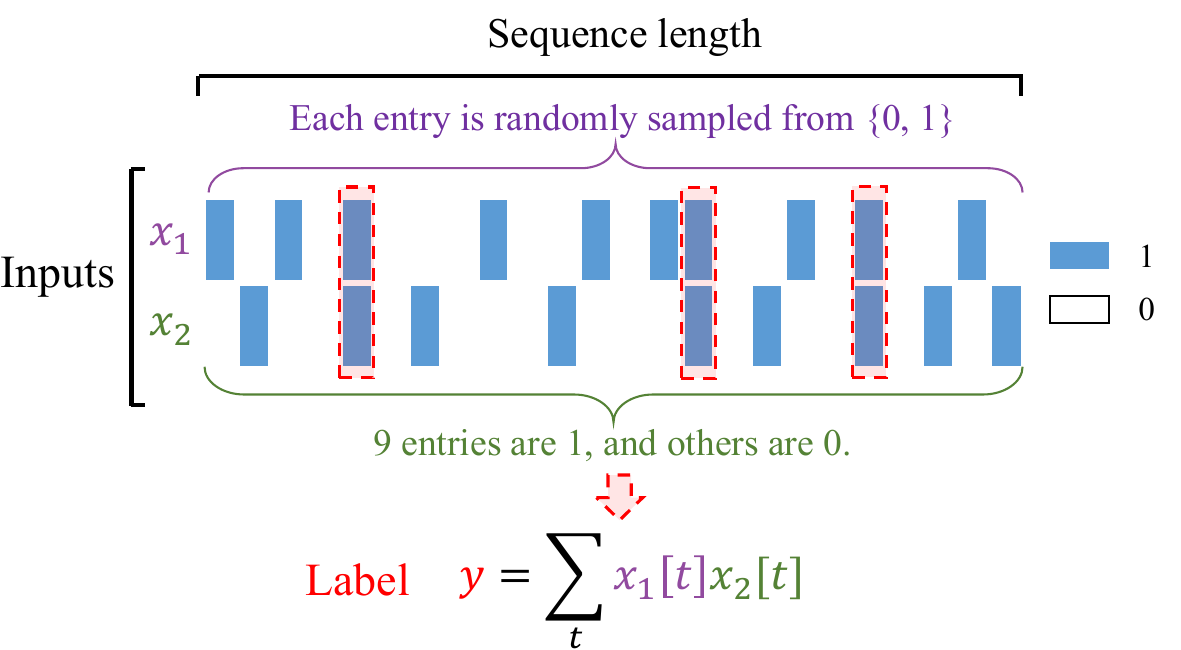}
    \caption{Illustration of the binary adding task, designed to test the ability of SNN models to capture long-range dependencies. The sequence length in this task is adjustable, allowing flexibility in controlling the task's difficulty.}
    \label{fig:binaryadd}
\end{figure}

In summary, our analysis utilizing the STP tool reveals that the current neuromorphic benchmarks are inadequate for evaluating the temporal processing capabilities of SNNs.

\section{Temporal Processing Benchmark Suite}
\label{sec:benchmark}
To bridge this gap, we present a new benchmark suite comprising three tasks in this section, along with a validation study to demonstrate their effectiveness in assessing temporal processing capabilities.

\subsection{Benchmark Suite}
To reveal the status quo of SNNs for temporal processing, we present a benchmark suite designed to comprehensively evaluate the temporal processing capabilities of existing SNN approaches. This suite incorporates three tasks with distinct modalities: language generation, pixel-level image classification, and mathematics.

\begin{itemize}
\item{\textbf{Penn Treebank (PTB)}. PTB is a widely used language modeling dataset \cite{marcus1993building}, derived from Wall Street Journal articles, including various text types such as news reports, editorials, and financial analyses. In this task, the text is tokenized into individual words with a vocabulary size of 10,000. Each sample is then truncated into sequences of 70 tokens. The model is required to predict the next token at each time step based on the preceding input context. }
\item{\textbf{Permuted-Sequential MNIST (PS-MNIST)}. PS-MNIST is a sequence classification dataset derived from the MNIST dataset~\cite{lecun1998gradient}. Each gray-scaled image from the MNIST dataset is first flattened in row-major order into a sequence of length 784. The pixel order is then shuffled using a fixed random permutation matrix, disrupting locality to challenge the model's ability to model long-range dependencies. Finally, these pixels are fed to an SNN one pixel at a time, with the prediction made at the final time step.} 

\item{\textbf{Binary Adding}. To challenge models in performing temporal processing over long distances, we propose a novel binary adding task. As illustrated in Fig.~\ref{fig:binaryadd}, we generate a synthetic $10$-class dataset with two input channels, denoted as $\{\boldsymbol{x}_1, \boldsymbol{x}_2\} \in \{0,1\}^ {T }$, where $\boldsymbol{x}_1$ represents a binary value sequence and $\boldsymbol{x}_2$ serves as an index indicator. The input sequence length $T$ can be flexibly adjusted to change the task difficulty. The binary value sequences $\boldsymbol{x}_1$ consist of $T$ entries randomly sampled from $\{0,1\}$, while the indicator sequence $\boldsymbol{x}_2$ assigns a value of $1$ to nine randomly selected indices and $0$ otherwise. The indicator sequence $\boldsymbol{x}_2$ acts as a pointer to indicate which entries in $\boldsymbol{x}_1$ should be added. The label is defined as $y=\sum_{t=1}^{T}{\boldsymbol{x}_1[t]\cdot\boldsymbol{x}_2[t]}$, which ranges from $0$ to $9$. The model must process the entire sequence before making a prediction, necessitating the ability to integrate information over a long time span. For a consistent comparison across different SNN approaches, we construct a fixed dataset containing $50,\!000$ training samples and $2,\!000$ testing samples.}
\end{itemize}

\subsection{Validation of Benchmarks using the STP}

\begin{figure}[!t]
\centering\includegraphics[width=0.84\linewidth]{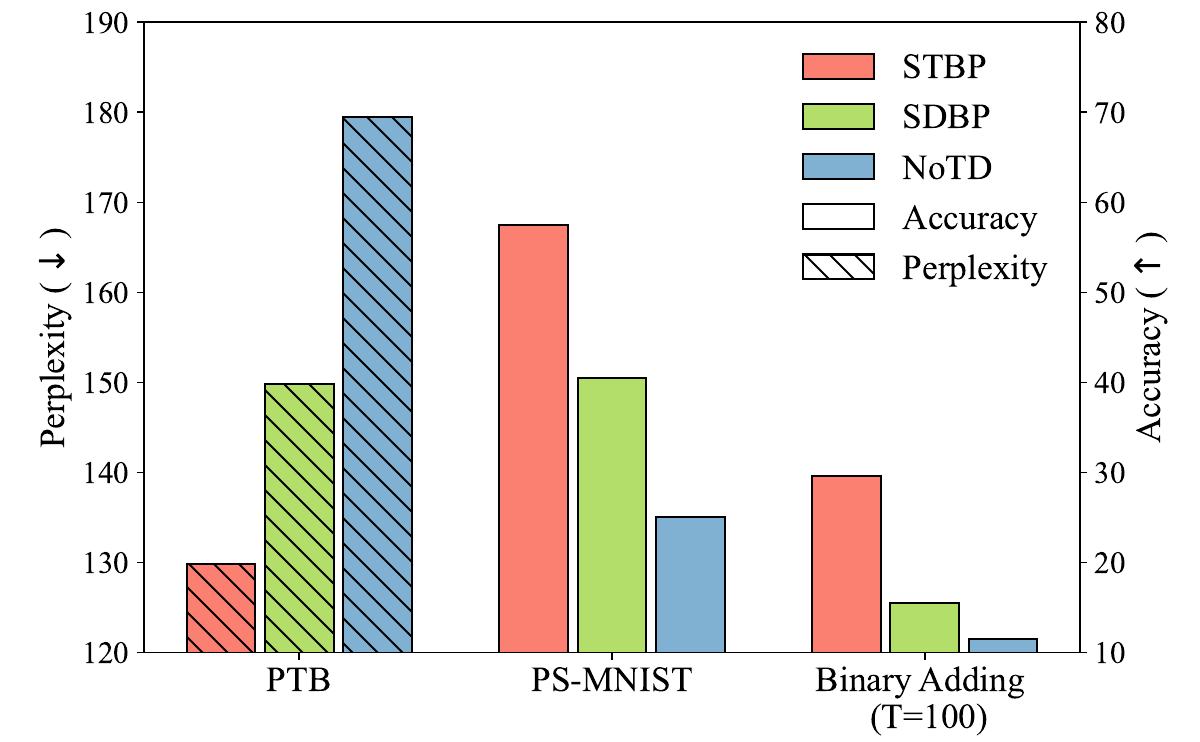}
\caption{Validation of the temporal processing benchmark suite through the STP}
\label{Fig:res_sequence_bench}
\end{figure}

We further apply the STP tool to validate the effectiveness of these three benchmarks in assessing temporal processing capabilities. Details of the training setups are provided in Appendix~\ref{app:imp_det_seq_bench}. As shown in Fig.~\ref{Fig:res_sequence_bench}, STBP significantly outperforms SDBP, which in turn substantially surpasses NoTD. This observation is consistent across all three benchmarks, demonstrating that these benchmarks contain critical temporal information that is necessary to be captured for high performance. Therefore, they can serve as effective benchmarks for evaluating the temporal processing capabilities of various SNN approaches.


\section{Status Quo of SNNs for temporal processing}
\label{sec:benchmarking}
To assess the current state of SNNs in temporal processing, we further conduct a comprehensive study of over thirty recently developed SNN methods using our proposed benchmark suite. The evaluated methods encompass a wide range of aspects, including learning algorithms in Section~\ref{subsec:training_algo}, surrogate gradients in Section~\ref{subsec:sg}, normalization techniques in Section~\ref{subsec:norm}, neuron models in Section~\ref{subsec:neuron_models}, and neural architectures in Section~\ref{subsec:network_architecture}. For completeness, each method is evaluated using both feedforward and recurrent architectures~\cite{bellec2018long,bellec2020solution}. The Recurrent SNNs (RSNNs) incorporate recurrent weights to retain historical information, thereby offering improved memory capacity compared to feedforward architectures.

\begin{table}[!t]
\centering
\caption{Comparison of learning algorithms on temporal processing benchmarks. ``FF" and ``Rec." refer to ``feedforward" and ``recurrent" architectures, respectively. ``PPL" stands for ``perplexity."}
\begin{tabular}{l|cc|cc|cc}
\hline
\textbf{Dataset} & \multicolumn{2}{c|}{\begin{tabular}[c]{@{}c@{}}PTB \\ (\(T=70\))\end{tabular}} & \multicolumn{2}{c|}{\begin{tabular}[c]{@{}c@{}}PS-MNIST \\ (\(T=784\))\end{tabular}}
& \multicolumn{2}{c}{\begin{tabular}[c]{@{}c@{}}Binary Adding \\ (\(T=100\))\end{tabular}} \\ \hline
\textbf{Metric} & \multicolumn{2}{c|}{PPL ~$\downarrow$} & \multicolumn{2}{c|}{Acc.~$\uparrow$} & \multicolumn{2}{c}{Acc.~$\uparrow$} \\ \hline
\textbf{Method} & FF & Rec. & FF & Rec. & FF & Rec. \\ \hline
STBP~\cite{wu2018spatio} & 129.96 & 111.96 & 57.45 & 72.97 & 29.60 & 53.35 \\
T-STBP & 137.8 & 120.58 & 53.00 & 71.03 & 23.00 & 51.50 \\
E-prop~\cite{bellec2020solution} & - & 125.54 & - & 52.88 & - & 50.85 \\
OTTT~\cite{xiao2022online} & 141.77 & - & 44.61 & - & 17.20 & - \\
SLTT~\cite{meng2023towards} & 149.86 & - & 40.53 & - & 15.50 & - \\ \hline
\end{tabular}
\label{tab:training_algos}
\end{table}

To ensure a fair comparison across these methods, we re-implement each method to the best of our ability, utilizing their publicly available source codes and descriptions provided in their respective papers. We maintain consistent training setups across all methods. To optimize the hyperparameters for each method, we conduct a grid search; detailed information on the hyperparameters and training setups is available in our open-sourced library. We have open-sourced our code and benchmarks to facilitate future advancements in the field, and we welcome contributions from the neuromorphic computing community to expand our benchmark by including more tasks and recent SNN methods.

\begin{figure}[!htb]
    \centering\includegraphics[width=0.9\linewidth]{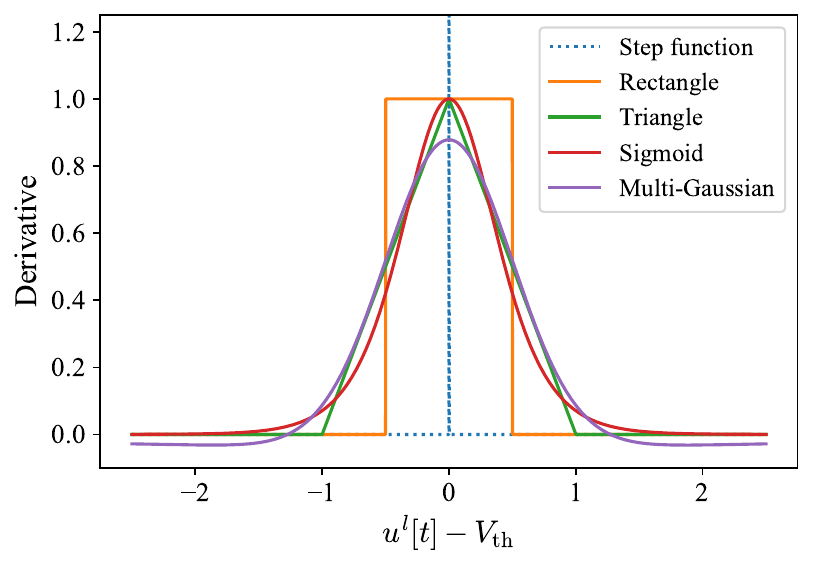}
    \caption{Comparison of different surrogate gradient functions.}
    \label{fig:sg}
\end{figure}

\subsection{Learning Algorithms}
\label{subsec:training_algo}
We begin by evaluating the performance of five SNN learning algorithms, including two offline learning algorithms (i.e., STBP~\cite{wu2018spatio}, Truncated-STBP~(T-STBP)) as well as three recently-proposed online learning algorithms (i.e., E-prop~\cite{bellec2020solution}, Online Training Through Time~(OTTT)~\cite{xiao2022online},  and Spatial Learning Through Time~(SLTT)~\cite{meng2023towards}). 
STBP is a standard gradient-based learning algorithm, which unfolds SNNs over both spatial and temporal domains and applies gradient descent to the entire computational graph. T-STBP processes input sequences in smaller segments, preventing error gradients from being backpropagated between these segments while preserving the complete temporal gradients within each individual segment. In contrast, the three online learning algorithms update SNNs at every time step, and they either partially or fully disregard temporal gradients to enable online learning. Specifically, E-prop achieves online learning in RSNNs by using eligibility traces, which accumulate presynaptic activities over time. Similarly, OTTT extends the mechanism of eligibility traces to feedforward SNNs, enabling effective online training for large-scale datasets. SLTT, on the other hand, ignores all temporal gradients, removing the need for additional traces while slightly compromising gradient precision.

In our evaluation, we adhere to the original setups for these algorithms, applying OTTT and SLTT exclusively to feedforward SNNs, while E-prop is used solely for RSNNs. Results presented in Table~\ref{tab:training_algos} indicate that STBP consistently outperforms the other four learning algorithms across all three benchmarks, regardless of the architecture. This performance gap is attributed to the omission of temporal gradients in these algorithms, which leads to biased gradient estimations and impedes the accurate propagation of temporal errors.

\begin{table}[!t]
\centering
\caption{Comparison of surrogate gradient functions on temporal processing benchmarks. The  \textbf{best} model is highlighted in \textbf{bold}, and the {\ul second} best is {\ul underlined}.}
\setlength{\tabcolsep}{1.5mm}
\begin{tabular}{l|cc|cc|cc}
\hline
\textbf{Dataset} & \multicolumn{2}{c|}{\begin{tabular}[c]{@{}c@{}}PTB \\ (\(T=70\))\end{tabular}} & \multicolumn{2}{c|}{\begin{tabular}[c]{@{}c@{}}PS-MNIST \\ (\(T=784\))\end{tabular}}
& \multicolumn{2}{c}{\begin{tabular}[c]{@{}c@{}}Binary Adding \\ (\(T=100\))\end{tabular}} \\  \hline
\textbf{Metric} & \multicolumn{2}{c|}{PPL~$\downarrow$} & \multicolumn{2}{c|}{Acc.~$\uparrow$} & \multicolumn{2}{c}{Acc.~$\uparrow$}  \\ \hline
\textbf{Method} & FF & Rec. & FF & Rec. & FF & Rec. \\ \hline
Rectangle~\cite{wu2018spatio} & 129.96 & 111.96 & 57.45 & 72.97 & {\ul 29.60} & 53.35 \\
Triangle~\cite{deng2022temporal} & {\ul 127.78} & {\ul 108.98} & {\ul 62.96} & \textbf{77.15} & 28.20 & {62.90} \\
Multi-Gaussian~\cite{ALIF} & 130.01 & 112.74 & 57.01 & 71.96 & \textbf{32.20} & {\ul 65.60} \\
Sigmoid~\cite{adLIF} & \textbf{127.33} & \textbf{107.92} & \textbf{63.18} & {\ul 76.02} & {29.35} & \textbf{66.45} \\
ASGL~\cite{ASGL} & 128.14 & 113.18 & 58.32 & 66.00 & 26.15 & 46.70 \\ \hline
\end{tabular}
\label{tab:sg}
\end{table}

Notably, our findings contrast with previous studies~\cite{xiao2022online,bellec2020solution,meng2023towards} that reported nearly lossless performance for these online learning algorithms compared to STBP. This discrepancy stems from the fact that earlier evaluations were primarily on datasets with no or limited temporal information, such as CIFAR10, CIFAR10-DVS, and DvsGesture, where the absence of temporal gradients had a minimal effect. In contrast, our benchmarks necessitate effective learning over time, where the accurate calculation of temporal gradients is crucial. This experiment underscores the limitations of commonly used neuromorphic benchmarks and highlights a significant accuracy gap between existing online learning methods and STBP in temporal processing tasks.

\subsection{Surrogate Gradient Functions}
\label{subsec:sg}

To address the nondifferentiable activation functions used in SNNs, surrogate gradient methods \cite{8891809} are commonly employed. These methods retain the original step function during the forward pass while replacing it with a smooth and continuous function during the backward pass. In this section, we evaluate the performance of five most frequently used surrogate gradient functions: Rectangle \cite{bengio2013estimating, wu2018spatio}, Triangle \cite{8891809,deng2022temporal}, Sigmoid \cite{adLIF}, Multi-Gaussian \cite{ALIF}, and Adaptive Smoothing Gradient Learning (ASGL) \cite{ASGL}. Notably, ASGL~\cite{ASGL} not only smooths the step function during backpropagation but also partially replaces the step function in the forward pass with the integral of surrogate functions during training. This approach mitigates the gradient approximation error caused by the mismatch between actual and surrogate gradients, resulting in smoother training.

\begin{table}[!t]
\centering
\caption{Comparison of normalization methods on temporal processing benchmarks}
\begin{tabular}{l|cc|cc|cc}
\hline
\textbf{Dataset} & \multicolumn{2}{c|}{\begin{tabular}[c]{@{}c@{}}PTB \\ (\(T=70\))\end{tabular}} & \multicolumn{2}{c|}{\begin{tabular}[c]{@{}c@{}}PS-MNIST \\ (\(T=784\))\end{tabular}}
& \multicolumn{2}{c}{\begin{tabular}[c]{@{}c@{}}Binary Adding \\ (\(T=100\))\end{tabular}} \\  \hline
\textbf{Metric} & \multicolumn{2}{c|}{PPL~$\downarrow$} & \multicolumn{2}{c|}{Acc.~$\uparrow$} & \multicolumn{2}{c}{Acc.~$\uparrow$} \\ \hline
\textbf{Method} & FF & Rec. & FF & Rec. & FF & Rec. \\ \hline
w/o BN & 129.96 & 111.96 & 57.45 & 72.97 & 29.60 & 53.35 \\
TEBN~\cite{deng2022temporal} & 126.88 & 102.24 & \textbf{94.94} & \textbf{95.02} & \textbf{53.03} & \textbf{65.45} \\
TDBN~\cite{zheng2021going} & 127.04 & 101.54 & 72.74 & 88.60 & 35.55 & 64.10 \\
LayerNorm~\cite{ba2016layer} & \textbf{123.54} & \textbf{101.53} & 62.28 & 69.96 & 40.25 & 49.70 \\ \hline
\end{tabular}
\label{tab:bns}
\end{table}

Results in Table \ref{tab:sg} indicate that while the performance of different surrogate functions shows only minor differences in feedforward architectures, their impact is significantly more pronounced in recurrent architectures. This is due to the recurrent connection from the output spike at time $t-1$ to the membrane potential at time $t$, which introduces an additional pathway for gradient backpropagation. This iterative computation of surrogate gradients over time introduces additional challenges for SNN training. Fortunately, a well-shaped surrogate function can more effectively propagate gradients back to earlier time steps through this recurrent pathway, resulting in improved network convergence for recurrent architectures.

To identify the most suitable surrogate gradient functions for temporal processing tasks, we further rank their performance across the three tasks and two architectures. The average ranking is used as a score to quantify their effectiveness. Our results indicate that the Sigmoid~\cite{adLIF} and Triangle~\cite{deng2022temporal} functions achieve the top two average rankings, with scores of $0.67$ and $1.17$, respectively. This finding suggests that these two functions are more appropriate for temporal tasks. The superior performance of the Sigmoid and Triangle functions can be attributed to two factors. Firstly, their higher smoothness facilitates error propagation during SNN training~\cite{wu2018spatio, deng2022temporal, eshraghian2023training}. Additionally, the values of these surrogate functions provide a closer approximation to the ill-defined derivative of the step function, as illustrated by the dotted curves in Fig.~\ref{fig:sg}. Consequently, the gradient mismatch problem \cite{ASGL} can be more effectively alleviated, contributing to better training convergence. Finally, we examine the performance of the ASGL strategy~\cite{ASGL}. Despite the performance improvements observed in previous static tasks, our results indicate that this strategy is less effective for temporal processing tasks. This inefficacy primarily arises from the discrepancy between the spike generation function employed during training and that used during inference, which accumulates over time, particularly when handling long sequences.

\subsection{Normalization Methods}
\label{subsec:norm}
Here, we evaluate the influence of normalization methods on the temporal processing performance of SNNs. Building upon the Batch Normalization (BN)~\cite{ioffe2015batch} method in ANNs, normalization methods have been developed specifically for SNNs to improve training stability and accelerate convergence. For example, Threshold-Dependent BN (TDBN)~\cite{zheng2021going} normalizes features across both batch and time dimensions and adjusts the normalized variance based on the threshold, which effectively mitigates the problems of gradient vanishing or explosion during training. Furthermore, Temporal Effective BN (TEBN)~\cite{duan2022temporal} incorporates trainable factors to rescale presynaptic inputs at each time step. This technique smooths temporal distributions of gradient norms and stabilizes training. In addition to TDBN and TEBN, we also assess Layer Normalization~(LayerNorm)~\cite{ba2016layer}, which is commonly used in non-spiking sequence models and can be seamlessly applied to SNNs.

Three major observations can be drawn from the results presented in Table~\ref{tab:bns}. First,  TEBN, TDBN, and LayerNorm all significantly enhance performance across the three benchmarks. Second, TEBN, in particular, shows substantial accuracy improvements on the PS-MNIST and binary adding tasks. The large improvement is attributed to the rescaling of presynaptic inputs with trainable factors at each time step in TEBN. This mechanism helps stabilize gradient flow over time and ensures that relevant information is preserved throughout the learning process. Third, LayerNorm proves to be particularly effective in language modeling tasks like PTB. Its ability to normalize across non-batch dimensions facilitates the learning of temporal sequences with varying effective lengths, leading to better performance in handling complex temporal data.

\begin{figure*}[!t]
\centering\includegraphics[width=0.8\linewidth]{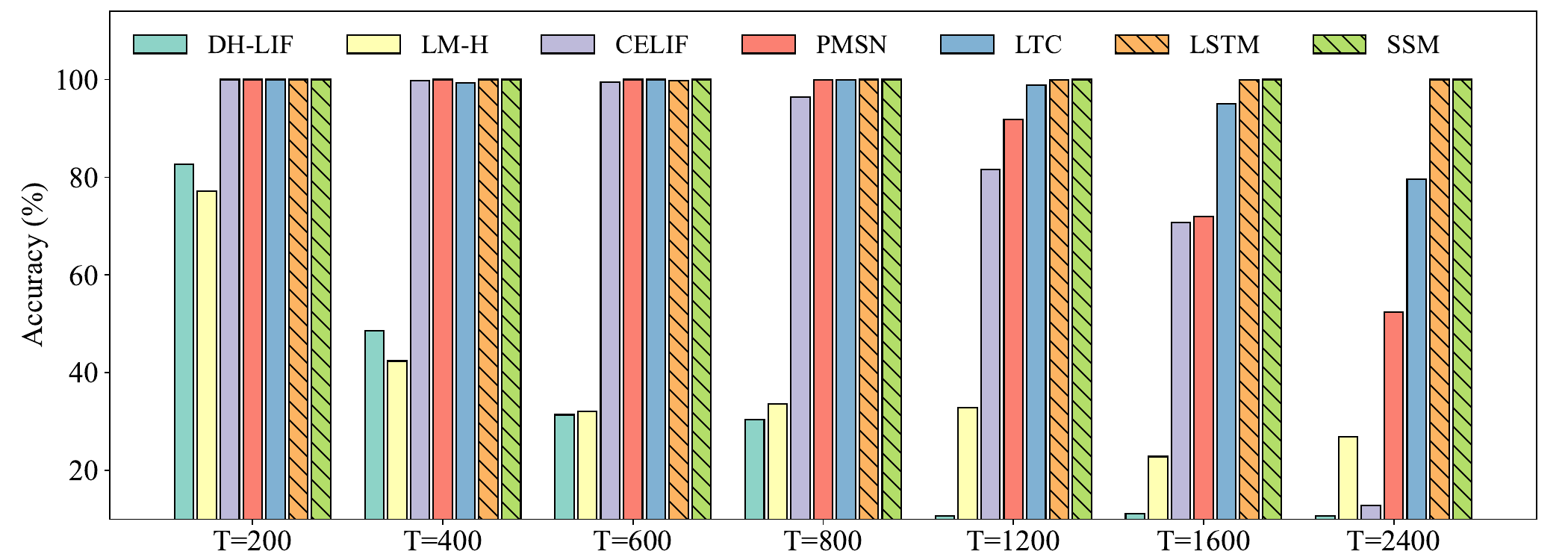}
\caption{Comparison of advanced spiking and non-spiking sequence models. Each model is evaluated on the binary adding task with sequence lengths ranging from $200$ to $2400$.
} 
\label{Fig:snnrnn}
\end{figure*}

\begin{table}[!t] 
\centering
\caption{Comparison of spiking neuron models on temporal processing benchmarks. The  \textbf{best} model is highlighted in \textbf{bold}, the {\ul second} best is {\ul underlined}, and the \textit{third} best is \textit{italicized}.}
\setlength{\tabcolsep}{1.5mm}
\begin{tabular}{l|cc|cc|cc}
\hline
\textbf{Dataset} & \multicolumn{2}{c|}{\begin{tabular}[c]{@{}c@{}}PTB \\ (\(T=70\))\end{tabular}} & \multicolumn{2}{c|}{\begin{tabular}[c]{@{}c@{}}PS-MNIST \\ (\(T=784\))\end{tabular}}
& \multicolumn{2}{c}{\begin{tabular}[c]{@{}c@{}}Binary Adding \\ (\(T=100\))\end{tabular}} \\  \hline
\textbf{Metric} & \multicolumn{2}{c|}{PPL~$\downarrow$} & \multicolumn{2}{c|}{Acc.~$\uparrow$} & \multicolumn{2}{c}{Acc.~$\uparrow$} \\ \hline
\textbf{Network} & FF & Rec. & FF & Rec. & FF & Rec. \\ \hline
\textbf{\#Params.} & $\sim$5M & $\sim$6M & $\sim$90K & $\sim$160K & $\sim$20K & $\sim$40K \\ \hline
LIF & 129.96 & 111.96 & 57.45 & 72.97 & 29.60 & 53.35 \\
PLIF
\cite{PLIF} & 123.76 & 105.64 & 55.86 & 77.32 & 29.40 & 53.25 \\
ALIF
\cite{ALIF} & 113.67 & 102.25 & 73.90 & 85.78 & 40.30 & 68.00 \\
adLIF
\cite{adLIF} & 118.52 & \textbf{97.22} & 85.93 & 89.53 & 42.00 & {99.05} \\
GLIF
\cite{GLIF} & {\ul 111.58} & 103.07 & \textit{95.42} & {\ul 95.04} & 90.15 & 63.60 \\
LTC
\cite{ltc} & \textbf{104.10} & {\ul 99.09} & 86.33 & 90.94 & \textbf{100.00} & \textbf{100.00} \\
SPSN
\cite{PSN} & 120.43 & - & 83.88 & - & 45.70 & - \\
TCLIF
\cite{TCLIF} & 286.71 & 255.67 & 86.81 & \textit{92.08} & 19.10 & 19.90 \\
LM-H
\cite{LMH} & 122.69 & 102.05 & 77.70 & 83.14 & \textit{99.25} & 96.10 \\
CLIF
\cite{CLIF} & 128.28 & 108.21 & 43.90 & 70.44 & 19.10 & 64.30 \\
DH-LIF
\cite{dhsnn} & 115.61 & \textit{100.55} & 79.12 & 91.07 & 98.85 & \textit{99.35} \\
CELIF
\cite{CELIF} & \textit{112.35} & 106.52 & \textbf{97.76} & \textbf{97.66} & 48.40 &\textbf{100.00} \\
PMSN
\cite{PMSN} & 113.24 & - & {\ul 96.28} & - & \textbf{100.00} & - \\ 
\hline
\end{tabular}
\label{tab:neuron}
\end{table}

\subsection{Spiking Neuron Models}
\label{subsec:neuron_models}

In this section, we conduct a comprehensive benchmarking of existing spiking neuron models for temporal processing. The widely used LIF model~\cite{lif} has limited memory capacity and often suffers from the temporal gradient vanishing problem~\cite{TCLIF}, which results in relatively poor temporal processing capabilities. Recently, several advanced spiking neuron models have been proposed to provide enhanced temporal modeling capabilities. These models can be categorized into two main types. The first category encompasses single-compartment spiking neuron models, which represent biological neurons as undivided units with enriched neuronal dynamics. For instance, Parametric LIF (PLIF) \cite{PLIF} introduces a learnable leaky constant for each neuron, allowing the model to adapt to diverse decaying rates of the input. Adaptive LIF (ALIF) \cite{ALIF} and Liquid Time-Constant (LTC) \cite{ltc} models incorporate the adaptive threshold as an additional state variable, allowing the spiking neuron to retain information related to its firing history within the firing threshold. Furthermore, models such as Generalized Leaky Integrate-and-Fire (GLIF) \cite{GLIF}, Complementary Leaky Integrate-and-Fire (CLIF) \cite{CLIF}, adaptive Leaky Integrate-and-Fire (adLIF) \cite{adLIF}, and Context Embedding Leaky Integrate-and-Fire (CELIF) \cite{CELIF} integrate various biological mechanisms such as ion channels, complementary membrane potentials, adaptation currents, and temporal context to enhance neuronal dynamics of spiking neurons.

In another vein of research, several multi-compartment spiking neuron models have also been proposed. Inspired by P-R neurons in the hippocampus, the Two-Compartment LIF (TC-LIF) model \cite{TCLIF} is specifically designed to model interactions between the soma and dendrites. To further reduce the parameter constraints of the TC-LIF model, the Learnable Multi-Hierarchical (LM-H) model \cite{LMH} has been proposed, demonstrating more stable gradient propagation in deep networks. Going beyond models with meticulously designed two compartments, the Dendritic Heterogeneity LIF (DH-LIF) model \cite{dhsnn} endows multiple dendritic compartments with heterogeneous decaying time constants. To address concerns about the slow training speed of SNNs, recent research has also proposed several neuron models to enable parallel computation in the temporal dimension. The family of Parallel Spiking Neural (PSN) models \cite{PSN} transforms the charging dynamics of the membrane potential into a learnable decay matrix and bypasses the neuron's reset mechanism to enable parallel computation. Furthermore, the Parallel Multi-Compartment Neuron (PMSN) model \cite{PMSN} incorporates multiple interconnected neuronal compartments. These inter-compartment interactions can effectively represent temporal information across diverse timescales.

Since most of these advanced models are developed based on static image datasets, their efficacy for temporal processing tasks remains elusive. To this end, we comprehensively compare their performance against the LIF model on our temporal processing benchmarks.  It is important to note that different studies have adopted different numbers of learnable model parameters for their respective neuron models. To ensure a fair comparison, we proportionally adjust the number of neurons in the hidden layers so that all neuron models are evaluated with the same number of parameters as the baseline. As the results presented in Table \ref{tab:neuron}, the various neuron models exhibit varying degrees of improvement over the LIF model, highlighting their effectiveness in enhancing the temporal processing capabilities of SNNs. 

The observed improvements can be explained by several specific architectural features of these models. For instance, the improved performance achieved by ALIF can be attributed to 
the additional slow-decaying state variables, which facilitate information integration over longer timespans. Moreover, the heterogeneous decaying rate of state variables used in these models can further facilitate the establishment of temporal dependencies across different timescales, as demonstrated in models like GLIF and DH-LIF. Furthermore, models such as adLIF, TCLIF, LM-H, and PMSN incorporate recurrent interactions between neuronal variables. These enriched neuronal dynamics significantly enhance the integration of temporal information. Other strategies, such as the extended temporal receptive field employed by SPSN, the temporal context in CELIF, and the liquid decaying rate in LTC, also demonstrate the enhancement of temporal processing capability.

The results also highlight the importance of the time decaying factors in facilitating temporal processing. For instance, the TC-LIF neuron omits these decay factors, leading to continuous accumulation of information in the two membrane potential variables. While this configuration works well for static datasets, it encounters challenges when targets change over time, as observed in tasks such as the PTB and binary adding. This necessitates a neuron model capable of rapidly forgetting past information to respond promptly to new inputs. By mitigating the excessive accumulation of historical inputs, these decay factors ensure that the spiking neuron remains responsive to new stimuli over time.

\subsection{Comparison of Spiking and Non-Spiking Sequence Models}
\label{subsec:annvssnn}
Having demonstrated the effectiveness of various advanced SNN models in temporal processing,
we further benchmark their performance by comparing them against leading non-spiking sequence models. In the binary adding problem, models such as LM-H, DH-LIF, and PMSN using a feedforward architecture, as well as LTC and CELIF employing a recurrent architecture, can successfully accomplish the task with a sequence length of $T\!=\!100$. These results indicate competitive performance relative to prominent non-spiking RNN models such as Long Short-Term Memory (LSTM)~\cite{LSTM} and SSM~\cite{SSM}. To further figure out whether current advanced SNN models already achieve similar to or even surpass the performance of these RNN models, we quantify their long-range temporal processing capacity by varying $T$ from 100 to $200$, $400$, $600$, $800$, $1200$, $1600$, and $2400$. 

As shown in Fig.~\ref{Fig:snnrnn}, the performance of LM-H and DH-LIF models degrades significantly when $T$ reaches $400$. The performance of CELIF model starts to degrade beyond $T\!=\!600$ and fails to learn any meaningful temporal information when $T\!=\!2400$. Similarly, both LTC and PMSN show varying degrees of degradation as the sequence length increases. In contrast, SSM and LSTM consistently achieve nearly $100\%$ accuracy, even at the challenging scenario when $T\!=\!2400$. Collectively, these observations highlight that a significant gap still exists between SNNs and ANNs in modeling long-range dependencies. 

\begin{table}[!t]
\begin{threeparttable}
\centering
\caption{Comparison of delay learning and neural architectures on temporal processing tasks. $T_{\text{in}}$ represents the internal time window of spiking neurons.
}
\setlength{\tabcolsep}{4.6pt}
\begin{tabular}{l|cc|cc|cc}
\hline
\textbf{Dataset} & \multicolumn{2}{c|}{\begin{tabular}[c]{@{}c@{}}PTB \\ (\(T=70\))\end{tabular}} & \multicolumn{2}{c|}{\begin{tabular}[c]{@{}c@{}}PS-MNIST \\ (\(T=784\))\end{tabular}} & \multicolumn{2}{c}{Binary Adding} \\ \hline
\textbf{Metric} & \multicolumn{2}{c|}{PPL~$\downarrow$} & \multicolumn{2}{c|}{Acc.~$\uparrow$} & \(T\)~$\uparrow$ & Acc.~$\uparrow$ \\ \hline
\textbf{\#Params.} & \multicolumn{2}{c|}{$\sim$5M} & \multicolumn{2}{c|}{$\sim$90K} & \multicolumn{2}{c}{$\sim$40K} \\ \hline
LIF & \multicolumn{2}{c|}{129.96} & \multicolumn{2}{c|}{57.45} & 100 & 34.15 \\
LIF w/ DCLS-Delays~\cite{DBLP:conf/iclr/HammouamriHM24} & \multicolumn{2}{c|}{89.87} & \multicolumn{2}{c|}{68.98} & 100 & 51.85 \\
\hline
TCN\tnote{*}\;\cite{TCN} & \multicolumn{2}{c|}{102.20} & \multicolumn{2}{c|}{95.10} & 1200 & 69.95 \\
SpikingTCN & \multicolumn{2}{c|}{114.46} & \multicolumn{2}{c|}{93.76} & 1200 & 61.95 \\ 
LSTM\tnote{*}\;\cite{LSTM} & \multicolumn{2}{c|}{88.08} & \multicolumn{2}{c|}{92.41} & 2400 & 100 \\
Gated Spiking Neuron~\cite{hao2024towards} & \multicolumn{2}{c|}{99.98} & \multicolumn{2}{c|}{80.13} & 1200 & 29.85 \\ 
Transformer\tnote{*}\;\cite{transformer} & \multicolumn{2}{c|}{112.43} & \multicolumn{2}{c|}{97.64} & 2400 & 100 \\
Spike-Driven Transformer\tnote{4}\;\cite{yao2024spike} & \multicolumn{2}{c|}{152.41} & \multicolumn{2}{c|}{96.21} & 2400 & 98.15 \\ 
Spike-Driven Transformer\tnote{1}\;\cite{yao2024spike} & \multicolumn{2}{c|}{327.82} & \multicolumn{2}{c|}{95.01} & 2400 & 88.05 \\ 
\hline
\end{tabular}
\begin{tablenotes}\footnotesize
\item[*]\!Non-spiking models.
\item[4, 1] $T_{\text{in}}=4$ and $T_{\text{in}}=1$, respectively.
\end{tablenotes}
\label{tab:architecture}
\end{threeparttable}
\end{table}

\subsection{Delay Learning and Neural Architectures}
\label{subsec:network_architecture}

In addition to improving the neuronal dynamics of spiking neurons, many recent studies have also investigated the enhancement of interactions among neurons. This section provides a comprehensive evaluation of these approaches, focusing specifically on the delay learning mechanism and neural architecture designs. 

Inspired by neuronal signaling in the brain, delay learning approaches incorporate axonal delay, the time it takes for signals to travel along an axon, into neuron modeling. In our experiments, we utilize a SOTA delay learning method, DCLS-Delays~\cite{DBLP:conf/iclr/HammouamriHM24}, which leverages 1-D temporal convolutions to enable effective delay modeling. The results in Table~\ref{tab:architecture} show that the LIF model combined with DCLS-Delays consistently outperforms its counterpart without delay modeling across all three benchmarks. This improvement underscores the significant advantage of delay learning, allowing the model to effectively establish temporal dependencies through the delay line. 

\begin{table}[!t]
\begin{threeparttable}
\centering
\caption{Comparison of energy efficiency between spiking architectures and their non-spiking counterparts on temporal processing tasks. The ratio is calculated as the energy cost of the non-spiking architecture divided by that of its spiking counterpart.
}
\setlength{\tabcolsep}{1.1pt}
\begin{tabular}{l|cc|cc|cc}
\hline
\textbf{Dataset} & \multicolumn{2}{c|}{\begin{tabular}[c]{@{}c@{}}PTB \\ (\(T=70\))\end{tabular}} & \multicolumn{2}{c|}{\begin{tabular}[c]{@{}c@{}}PS-MNIST \\ (\(T=784\))\end{tabular}} & \multicolumn{2}{c}{\makecell[c]{Binary Adding\\$(T=1200)$}} \\ \hline
\textbf{Energy} (\SI{}{\nano\joule}) & Cost & Ratio & Cost  & Ratio  & Cost  & Ratio  \\ 
\hline
TCN\tnote{*}\;\cite{TCN} &6535.7  & - &302.8  & - &187.4  & - \\
SpikingTCN  & 1505.6 & 4.3 &13.4 & 22.61 &6.8 & 27.56 \\ 
LSTM\tnote{*}\;\cite{LSTM} &8244.8 & - & 393.6& - &158.4 & -   \\
Gated Spiking Neuron~\cite{hao2024towards} & 1038.7 & 7.9 & 21.6 & 18.2 & 9.6 & 16.5 \\
Transformer\tnote{*}\;\cite{transformer}  &10223.0 & - & 1420.2 & - & 1059.8 & -  \\
Spike-Driven Transformer\tnote{4}\;\cite{yao2024spike} &1826.3 & 5.6 &131.3 & 10.8 &109.2 &9.7  \\
Spike-Driven Transformer\tnote{1}\;\cite{yao2024spike} & 367.3 & 27.8 &30.0 & 47.3 &23.6 & 44.9 \\ 
\hline
\end{tabular}
\begin{tablenotes}\footnotesize
\item[*]\!Non-spiking models.
\item[4,1] $T_{\text{in}}=4$ and $T_{\text{in}}=1$, respectively.
\end{tablenotes}
\label{tab:energy}
\end{threeparttable}
\end{table}

In terms of neural architectures, 
some recent studies focus on adapting advanced non-spiking architectures into spiking counterparts~\cite{yao2024spike,hao2024towards}. Here, we compare three well-established neural architectures for temporal processing: LSTM~\cite{LSTM}, TCN~\cite{TCN}, and Transformer~\cite{transformer}, with their spiking variants -- Gated Spiking Neuron (GSN)~\cite{hao2024towards}, SpikingTCN, and Spike-Driven Transformer~\cite{yao2024spike}. LSTM employs gating mechanisms to dynamically control information storage and forgetting, which effectively alleviates the problems of gradient vanishing and exploding. Similarly, GSN also adopts the gating mechanism to control the storage and forgetting of historical information. Unlike LSTMs, which process temporal sequences through iterative updates, TCNs use dilated convolutions to efficiently capture long-range dependencies. Recently, Transformer architectures have transformed temporal processing through the self-attention mechanism. This architecture demonstrates superior capability in managing long-range temporal dependencies, along with enhanced temporal parallelism and scalability. 

For Transformer, its spiking variant replaces the original continuous activation functions with the discrete step function employed in spiking neurons. In addition, an internal time window~$T_{\text{in}}$ of the spiking neuron is utilized to calculate the firing rate, thereby expanding the representation space of spiking neurons. However, incorporating this extra time window incurs additional computational overhead. To evaluate its impact, we compare the performance of the Spike-Driven Transformer model with  $T_{\text{in}}\!=\!4$ and $T_{\text{in}}\!=\!1$. To ensure a fair comparison, all architectures are configured with a comparable number of parameters for the same task. For the binary adding task, the sequence length is gradually increased from \(100\) to \(2400\) until the architecture no longer performs. Both the maximum sequence length and accuracy are reported in Table~\ref{tab:architecture}.

A key observation from Table~\ref{tab:architecture} is that, although the performance of spiking architectures is generally inferior to that of their non-spiking counterparts, the performance gap is relatively small, particularly when $T_{\text{in}}\!=\!4$. While the GSN architecture exhibits a significant accuracy gap compared to LSTM, it outperforms many of the advanced spiking neuron models presented in Table~\ref{tab:neuron}. Notably, the Spike-Driven Transformer excels in the binary adding task, underscoring its strong capability to model long-range dependencies through self-attention. The relatively lower performance of Spike-Driven Transformer on PTB is likely due to the overfitting with the small dataset size, and it is anticipated that performance would improve on a larger dataset. Furthermore, the Spike-Driven Transformer experiences a significant drop in accuracy when $T_{\text{in}}\!=\!1$. This finding underscores the importance of utilizing an additional internal time window to enhance the representational capacity of spiking neurons, which is critical for bridging the performance gap with SOTA non-spiking sequence models. 

We further compare the energy efficiency of these spiking architectures with that of their non-spiking counterparts. The details of the energy consumption calculation are provided in Appendix \ref{sec:energy_·cost}. As shown in Table \ref{tab:energy}, spiking architectures are generally an order of magnitude more energy efficient than their non-spiking counterparts. This efficiency stems from the spike-based computation, which relies on efficient Accumulate (AC) operations rather than the more expensive Multiply-Accumulate (MAC) operations typically used in non-spiking architectures. Additionally, Spike-Driven Transformer with $T_{\text{in}}\!=\!4$ consumes approximately four times as much energy as the model with $T_{\text{in}}\!=\!1$. This indicates that the accuracy improvements afforded by the extra time window come at the expense of significantly increased energy consumption, potentially diminishing the competitiveness of SNNs. This trade-off warrants further investigation. Nonetheless, even with $T_{\text{in}}\!=\!4$, Spike-Driven Transformer still improves energy efficiency by dozens of times compared to its non-spiking counterpart. These results underscore the significant energy-saving benefit of SNNs, making them particularly advantageous for applications in energy-constrained environments.


\section{Challenges and Future Directions}
\label{sec:challenges}

In this section, we identify four key challenges in current research on SNNs for temporal processing and propose some future directions to overcome these challenges. The primary challenge arises from the inadequacy of current neuromorphic benchmarks in evaluating SNNs' temporal processing capabilities. This limitation hinders the advancement of SNNs in effectively tackling tasks that involve complex and long-range temporal dependencies. To overcome this challenge, future research should prioritize the development of comprehensive neuromorphic benchmarks specifically designed to evaluate the SNN's abilities to capture long-range temporal dependencies. To ensure practical relevance, these benchmarks should encompass a diverse array of real-world temporal processing tasks that necessitate the maintenance of temporal context over extended durations. Additionally, they should be designed to emphasize the strengths of neuromorphic computing, such as high energy efficiency and low latency. By establishing these benchmarks, we can attain a more accurate assessment of SNN's performance in temporal processing, thereby facilitating their application in real-world temporal signal processing scenarios.

Another major challenge arises from the significant accuracy drops in online learning algorithms when applied to temporal processing tasks. While these algorithms have effectively enhanced the training efficiency and adaptability of SNNs, they often compromise the precision of temporal gradients, leading to suboptimal performance in learning long-range temporal dependencies. Addressing this challenge requires the development of a new generation of online or on-chip learning algorithms that can effectively and efficiently learn long-range temporal dependencies. This advancement would enable SNNs to continuously adapt to diverse and dynamic real-world scenarios.

The third challenge is that existing SNN models struggle to model long-range dependencies. The primary issues likely arise from optimization challenges, such as the vanishing gradient problem, where gradients associated with earlier time steps become exponentially smaller, resulting in a bias toward short-term dependencies. Additionally, current SNNs face difficulties in effectively storing and retrieving information over time, which hinders SNNs' capabilities in temporal processing. To address this challenge, future research should draw inspiration from the structures and functions of biological neural networks to create spiking neuron models and neural architectures with enhanced memory storage and information retrieval capabilities.

The last challenge concerns the prohibitive training time of SNNs for long sequences. Due to the inherently temporal nature of SNNs, training times in earlier SNN studies scale linearly with sequence length, especially for spiking neuron models that involve non-linear dynamics. This scaling makes it challenging to fully exploit the parallel processing capabilities of hardware accelerators like GPUs. Consequently, training SNNs can be excessively time-consuming, which limits researchers' ability to rapidly validate new ideas and iteratively refine SNN approaches. Despite recent advancements in temporal parallel spiking neuron models that significantly accelerate training, their reliance on linear recurrency as a prerequisite for parallelism fundamentally constrains their ability to capture the rich nonlinear dynamics present in biological neurons. Future work should focus on developing novel spiking neuron models that facilitate parallelized training over time, particularly by supporting the linearization of complex dynamics. Such advancements would drastically reduce training times, thereby enabling rapid development of SNN approaches.

\section{Conclusion}
\label{sec:conclusion}
In this work, we introduce the STP, a novel analytical tool designed to evaluate the effectiveness of benchmarks in assessing the temporal processing capabilities of SNN. Our application of the STP to widely used neuromorphic benchmarks indicates that these benchmarks can often be addressed without modeling the temporal dependencies of distant inputs. This finding suggests a significant deficiency in temporal information within current benchmarks, rendering them inadequate for a comprehensive evaluation of various methods' temporal processing capabilities. 

To further elucidate the current state of SNN approaches in temporal processing, we developed a suite of benchmarks encompassing three distinct temporal processing tasks and conducted a thorough comparison of existing SNN methodologies. Our benchmarking study demonstrated the enhanced temporal processing capabilities of recently introduced spiking neuron models and architectures. However, it also revealed that, despite the significant energy efficiency of SNNs, there remains a notable performance gap compared to SOTA non-spiking sequence models, particularly in the context of long sequences. Moreover, we discuss the challenges and future directions for SNNs in temporal processing, emphasizing the urgent need for the development of specialized neuromorphic benchmarks, learning algorithms, and computational models specifically tailored to real-world temporal processing tasks.

\bibliographystyle{IEEEtran}
\bibliography{myRefs}

\begin{thebibliography}{10}
\providecommand{\url}[1]{#1}
\csname url@samestyle\endcsname
\providecommand{\newblock}{\relax}
\providecommand{\bibinfo}[2]{#2}
\providecommand{\BIBentrySTDinterwordspacing}{\spaceskip=0pt\relax}
\providecommand{\BIBentryALTinterwordstretchfactor}{4}
\providecommand{\BIBentryALTinterwordspacing}{\spaceskip=\fontdimen2\font plus
\BIBentryALTinterwordstretchfactor\fontdimen3\font minus \fontdimen4\font\relax}
\providecommand{\BIBforeignlanguage}[2]{{%
\expandafter\ifx\csname l@#1\endcsname\relax
\typeout{** WARNING: IEEEtran.bst: No hyphenation pattern has been}%
\typeout{** loaded for the language `#1'. Using the pattern for}%
\typeout{** the default language instead.}%
\else
\language=\csname l@#1\endcsname
\fi
#2}}
\providecommand{\BIBdecl}{\relax}
\BIBdecl

\bibitem{RNN}
M.~Schuster and K.~K. Paliwal, ``Bidirectional recurrent neural networks,'' \emph{IEEE transactions on Signal Processing}, vol.~45, no.~11, pp. 2673--2681, 1997.

\bibitem{LSTM}
S.~Hochreiter and J.~Schmidhuber, ``Long short-term memory,'' \emph{Neural Computation}, vol.~9, no.~8, pp. 1735--1780, November 1997.

\bibitem{TCN}
S.~Bai, J.~Z. Kolter, and V.~Koltun, ``An empirical evaluation of generic convolutional and recurrent networks for sequence modeling,'' \emph{CoRR}, vol. abs/1803.01271, 2018.

\bibitem{transformer}
A.~Vaswani, N.~Shazeer, N.~Parmar, J.~Uszkoreit, L.~Jones, A.~N. Gomez, L.~u. Kaiser, and I.~Polosukhin, ``Attention is all you need,'' in \emph{Advances in Neural Information Processing Systems}, vol.~30.\hskip 1em plus 0.5em minus 0.4em\relax Curran Associates, Inc., 2017.

\bibitem{SSM}
A.~Gu, K.~Goel, and C.~Re, ``Efficiently modeling long sequences with structured state spaces,'' in \emph{International Conference on Learning Representations}, 2022.

\bibitem{9043731}
L.~Deng, G.~Li, S.~Han, L.~Shi, and Y.~Xie, ``Model compression and hardware acceleration for neural networks: A comprehensive survey,'' \emph{Proceedings of the IEEE}, vol. 108, no.~4, pp. 485--532, 2020.

\bibitem{maass1997networks}
W.~Maass, ``Networks of spiking neurons: The third generation of neural network models,'' \emph{Neural Networks}, vol.~10, no.~9, pp. 1659--1671, 1997.

\bibitem{10636118}
G.~Li, L.~Deng, H.~Tang, G.~Pan, Y.~Tian, K.~Roy, and W.~Maass, ``Brain-inspired computing: A systematic survey and future trends,'' \emph{Proceedings of the IEEE}, vol. 112, no.~6, pp. 544--584, 2024.

\bibitem{roy2019towards}
K.~Roy, A.~Jaiswal, and P.~Panda, ``Towards spike-based machine intelligence with neuromorphic computing,'' \emph{Nature}, vol. 575, no. 7784, pp. 607--617, 2019.

\bibitem{rapid2013}
Q.~Yu, H.~Tang, K.~C. Tan, and H.~Li, ``Rapid feedforward computation by temporal encoding and learning with spiking neurons,'' \emph{IEEE Transactions on Neural Networks and Learning Systems}, vol.~24, no.~10, pp. 1539--1552, 2013.

\bibitem{davies2018loihi}
M.~Davies, N.~Srinivasa, T.~Lin, G.~N. Chinya, Y.~Cao, S.~H. Choday, G.~D. Dimou, P.~Joshi, N.~Imam, S.~Jain, Y.~Liao, C.~Lin, A.~Lines, R.~Liu, D.~Mathaikutty, S.~McCoy, A.~Paul, J.~Tse, G.~Venkataramanan, Y.~Weng, A.~Wild, Y.~Yang, and H.~Wang, ``Loihi: {A} neuromorphic manycore processor with on-chip learning,'' \emph{{IEEE} Micro}, vol.~38, no.~1, pp. 82--99, 2018.

\bibitem{pei2019towards}
J.~Pei, L.~Deng, S.~Song, M.~Zhao, Y.~Zhang, S.~Wu, G.~Wang, Z.~Zou, Z.~Wu, W.~He, F.~Chen, N.~Deng, S.~Wu, Y.~Wang, Y.~Wu, Z.~Yang, C.~Ma, G.~Li, W.~Han, H.~Li, H.~Wu, R.~Zhao, Y.~Xie, and L.~Shi, ``Towards artificial general intelligence with hybrid {Tianjic} chip architecture,'' \emph{Nature}, vol. 572, no. 7767, pp. 106--111, Aug. 2019.

\bibitem{ma2017darwin}
D.~Ma, J.~Shen, Z.~Gu, M.~Zhang, X.~Zhu, X.~Xu, Q.~Xu, Y.~Shen, and G.~Pan, ``Darwin: A neuromorphic hardware co-processor based on spiking neural networks,'' \emph{Journal of Systems Architecture}, vol.~77, pp. 43--51, 2017.

\bibitem{Darwin3}
D.~Ma, X.~Jin, S.~Sun, Y.~Li, X.~Wu, Y.~Hu, F.~Yang, H.~Tang, X.~Zhu, P.~Lin, and G.~Pan, ``Darwin3: A large-scale neuromorphic chip with a novel {ISA} and on-chip learning,'' \emph{National Science Review}, vol.~11, no.~5, p. nwae102, 03 2024.

\bibitem{yao2024nc}
M.~Yao, O.~Richter, G.~Zhao, N.~Qiao, Y.~Xing, D.~Wang, T.~Hu, W.~Fang, T.~Demirci, M.~De~Marchi, L.~Deng, T.~Yan, C.~Nielsen, S.~Sheik, C.~Wu, Y.~Tian, B.~Xu, and G.~Li, ``Spike-based dynamic computing with asynchronous sensing-computing neuromorphic chip,'' \emph{Nature Communications}, vol.~15, no.~1, p. 4464, May 2024.

\bibitem{10.1162/neco_a_01245}
X.~Ju, B.~Fang, R.~Yan, X.~Xu, and H.~Tang, ``An {FPGA} implementation of deep spiking neural networks for low-power and fast classification,'' \emph{Neural Computation}, vol.~32, no.~1, pp. 182--204, 01 2020.

\bibitem{herz2006modeling}
A.~V. Herz, T.~Gollisch, C.~K. Machens, and D.~Jaeger, ``Modeling single-neuron dynamics and computations: A balance of detail and abstraction,'' \emph{Science}, vol. 314, no. 5796, pp. 80--85, 2006.

\bibitem{wu2020deep}
J.~Wu, E.~Y{\i}lmaz, M.~Zhang, H.~Li, and K.~C. Tan, ``Deep spiking neural networks for large vocabulary automatic speech recognition,'' \emph{Frontiers in Neuroscience}, vol.~14, p. 199, 2020.

\bibitem{wu2018spiking}
J.~Wu, Y.~Chua, M.~Zhang, H.~Li, and K.~C. Tan, ``A spiking neural network framework for robust sound classification,'' \emph{Frontiers in Neuroscience}, vol.~12, p. 836, 2018.

\bibitem{sequence2016}
Q.~Yu, R.~Yan, H.~Tang, K.~C. Tan, and H.~Li, ``A spiking neural network system for robust sequence recognition,'' \emph{IEEE Transactions on Neural Networks and Learning Systems}, vol.~27, no.~3, pp. 621--635, 2016.

\bibitem{ptl}
J.~Wu, C.~Xu, X.~Han, D.~Zhou, M.~Zhang, H.~Li, and K.~C. Tan, ``Progressive tandem learning for pattern recognition with deep spiking neural networks,'' \emph{IEEE Transactions on Pattern Analysis and Machine Intelligence}, vol.~44, no.~11, pp. 7824--7840, 2022.

\bibitem{NEURIPS2022_523caec7}
Q.~Yang, J.~Wu, M.~Zhang, Y.~Chua, X.~Wang, and H.~Li, ``Training spiking neural networks with local tandem learning,'' in \emph{Advances in Neural Information Processing Systems}, vol.~35.\hskip 1em plus 0.5em minus 0.4em\relax Curran Associates, Inc., 2022, pp. 12\,662--12\,676.

\bibitem{10254579}
Y.~Hu, Q.~Zheng, X.~Jiang, and G.~Pan, ``{Fast-SNN}: Fast spiking neural network by converting quantized {ANN},'' \emph{IEEE Transactions on Pattern Analysis and Machine Intelligence}, vol.~45, no.~12, pp. 14\,546--14\,562, 2023.

\bibitem{Xu_2023_CVPR}
Q.~Xu, Y.~Li, J.~Shen, J.~K. Liu, H.~Tang, and G.~Pan, ``Constructing deep spiking neural networks from artificial neural networks with knowledge distillation,'' in \emph{Proceedings of the IEEE/CVF Conference on Computer Vision and Pattern Recognition (CVPR)}, June 2023, pp. 7886--7895.

\bibitem{deng2022temporal}
S.~Deng, Y.~Li, S.~Zhang, and S.~Gu, ``Temporal efficient training of spiking neural network via gradient re-weighting,'' in \emph{International Conference on Learning Representations}, 2022, pp. 1--14.

\bibitem{imloss}
Y.~Guo, Y.~Chen, L.~Zhang, X.~Liu, Y.~Wang, X.~Huang, and Z.~Ma, ``{IM-Loss}: Information maximization loss for spiking neural networks,'' in \emph{Advances in Neural Information Processing Systems}, vol.~35.\hskip 1em plus 0.5em minus 0.4em\relax Curran Associates, Inc., 2022, pp. 156--166.

\bibitem{ijcai2023p335}
S.~Lian, J.~Shen, Q.~Liu, Z.~Wang, R.~Yan, and H.~Tang, ``Learnable surrogate gradient for direct training spiking neural networks,'' in \emph{Proceedings of the Thirty-Second International Joint Conference on Artificial Intelligence, {IJCAI-23}}.\hskip 1em plus 0.5em minus 0.4em\relax International Joint Conferences on Artificial Intelligence Organization, 8 2023, pp. 3002--3010, main Track.

\bibitem{zhu2024online}
Y.~Zhu, J.~Ding, T.~Huang, X.~Xie, and Z.~Yu, ``Online stabilization of spiking neural networks,'' in \emph{The Twelfth International Conference on Learning Representations}, 2024.

\bibitem{jiang2024ndot}
H.~Jiang, G.~D. Masi, H.~Xiong, and B.~Gu, ``{NDOT}: Neuronal dynamics-based online training for spiking neural networks,'' in \emph{Forty-first International Conference on Machine Learning}, 2024.

\bibitem{shen2024rethinking}
H.~Shen, Q.~Zheng, H.~Wang, and G.~Pan, ``Rethinking the membrane dynamics and optimization objectives of spiking neural networks,'' in \emph{The Thirty-eighth Annual Conference on Neural Information Processing Systems}, 2024.

\bibitem{hybridcoding}
X.~Chen, Q.~Yang, J.~Wu, H.~Li, and K.~C. Tan, ``A hybrid neural coding approach for pattern recognition with spiking neural networks,'' \emph{IEEE Transactions on Pattern Analysis and Machine Intelligence}, vol.~46, no.~5, pp. 3064--3078, 2024.

\bibitem{attentionsnns}
M.~Yao, G.~Zhao, H.~Zhang, Y.~Hu, L.~Deng, Y.~Tian, B.~Xu, and G.~Li, ``Attention spiking neural networks,'' \emph{IEEE Transactions on Pattern Analysis and Machine Intelligence}, vol.~45, no.~8, pp. 9393--9410, 2023.

\bibitem{NEURIPS2023_b8734840}
Q.~Xu, Y.~Gao, J.~Shen, Y.~Li, X.~Ran, H.~Tang, and G.~Pan, ``Enhancing adaptive history reserving by spiking convolutional block attention module in recurrent neural networks,'' in \emph{Advances in Neural Information Processing Systems}, vol.~36.\hskip 1em plus 0.5em minus 0.4em\relax Curran Associates, Inc., 2023, pp. 58\,890--58\,901.

\bibitem{9328792}
Q.~Yu, S.~Song, C.~Ma, L.~Pan, and K.~C. Tan, ``Synaptic learning with augmented spikes,'' \emph{IEEE Transactions on Neural Networks and Learning Systems}, vol.~33, no.~3, pp. 1134--1146, 2022.

\bibitem{ternarySpike2024}
Y.~Guo, Y.~Chen, X.~Liu, W.~Peng, Y.~Zhang, X.~Huang, and Z.~Ma, ``Ternary spike: Learning ternary spikes for spiking neural networks,'' \emph{Proceedings of the AAAI Conference on Artificial Intelligence}, vol.~38, no.~11, pp. 12\,244--12\,252, Mar. 2024.

\bibitem{xing2024spikelm}
X.~Xing, Z.~Zhang, Z.~Ni, S.~Xiao, Y.~Ju, S.~Fan, Y.~Wang, J.~Zhang, and G.~Li, ``Spike{LM}: Towards general spike-driven language modeling via elastic bi-spiking mechanisms,'' in \emph{Forty-first International Conference on Machine Learning}, 2024.

\bibitem{hu2024toward}
Y.~Hu, Q.~Zheng, G.~Li, H.~Tang, and G.~Pan, ``Toward large-scale spiking neural networks: {A} comprehensive survey and future directions,'' \emph{CoRR}, vol. abs/2409.02111, 2024.

\bibitem{lecun1998gradient}
Y.~LeCun, L.~Bottou, Y.~Bengio, and P.~Haffner, ``Gradient-based learning applied to document recognition,'' \emph{Proceedings of the IEEE}, vol.~86, no.~11, pp. 2278--2324, 1998.

\bibitem{krizhevsky2009learning}
A.~Krizhevsky and G.~Hinton, ``Learning multiple layers of features from tiny images,'' University of Toronto, Toronto, Ontario, Tech. Rep.~0, 2009.

\bibitem{eshraghian2023training}
J.~K. Eshraghian, M.~Ward, E.~O. Neftci, X.~Wang, G.~Lenz, G.~Dwivedi, M.~Bennamoun, D.~S. Jeong, and W.~D. Lu, ``Training spiking neural networks using lessons from deep learning,'' \emph{Proceedings of the IEEE}, 2023.

\bibitem{orchard2015converting}
G.~Orchard, A.~Jayawant, G.~K. Cohen, and N.~Thakor, ``Converting static image datasets to spiking neuromorphic datasets using saccades,'' \emph{Frontiers in Neuroscience}, vol.~9, p. 437, 2015.

\bibitem{li2017cifar10}
H.~Li, H.~Liu, X.~Ji, G.~Li, and L.~Shi, ``{CIFAR10-DVS}: An event-stream dataset for object classification,'' \emph{Frontiers in Neuroscience}, vol.~11, p. 244131, 2017.

\bibitem{amir2017low}
A.~Amir, B.~Taba, D.~J. Berg, T.~Melano, J.~L. McKinstry, C.~di~Nolfo, T.~K. Nayak, A.~Andreopoulos, G.~Garreau, M.~Mendoza, J.~Kusnitz, M.~DeBole, S.~K. Esser, T.~Delbr{\"{u}}ck, M.~Flickner, and D.~S. Modha, ``A low power, fully event-based gesture recognition system,'' in \emph{{IEEE} Conference on Computer Vision and Pattern Recognition, {CVPR}}.\hskip 1em plus 0.5em minus 0.4em\relax {IEEE} Computer Society, 2017, pp. 7388--7397.

\bibitem{zhou2024enhancing}
S.~Zhou, B.~Yang, M.~Yuan, R.~Jiang, R.~Yan, G.~Pan, and H.~Tang, ``Enhancing {SNN}-based spatio-temporal learning: A benchmark dataset and cross-modality attention model,'' \emph{Neural Networks}, vol. 180, p. 106677, 2024.

\bibitem{warden2018speech}
P.~Warden, ``Speech commands: {A} dataset for limited-vocabulary speech recognition,'' \emph{CoRR}, vol. abs/1804.03209, 2018.

\bibitem{TIMIT}
J.~S. {Garofolo}, L.~F. {Lamel}, W.~M. {Fisher}, J.~G. {Fiscus}, and D.~S. {Pallett}, ``{{DARPA} {TIMIT} acoustic-phonetic continuous speech corpus {CD-ROM. NIST} speech disc 1-1.1},'' p. 27403, 1993.

\bibitem{pan2020efficient}
Z.~Pan, Y.~Chua, J.~Wu, M.~Zhang, H.~Li, and E.~Ambikairajah, ``An efficient and perceptually motivated auditory neural encoding and decoding algorithm for spiking neural networks,'' \emph{Frontiers in Neuroscience}, vol.~13, p. 1420, 2020.

\bibitem{cramer2020heidelberg}
B.~Cramer, Y.~Stradmann, J.~Schemmel, and F.~Zenke, ``The heidelberg spiking data sets for the systematic evaluation of spiking neural networks,'' \emph{IEEE Transactions on Neural Networks and Learning Systems}, vol.~33, no.~7, pp. 2744--2757, 2020.

\bibitem{wu2018spatio}
Y.~Wu, L.~Deng, G.~Li, J.~Zhu, and L.~Shi, ``Spatio-temporal backpropagation for training high-performance spiking neural networks,'' \emph{Frontiers in Neuroscience}, vol.~12, p. 331, 2018.

\bibitem{xiao2022online}
M.~Xiao, Q.~Meng, Z.~Zhang, D.~He, and Z.~Lin, ``Online training through time for spiking neural networks,'' \emph{Advances in Neural Information Processing Systems}, vol.~35, pp. 20\,717--20\,730, 2022.

\bibitem{meng2023towards}
Q.~Meng, M.~Xiao, S.~Yan, Y.~Wang, Z.~Lin, and Z.-Q. Luo, ``Towards memory-and time-efficient backpropagation for training spiking neural networks,'' in \emph{Proceedings of the IEEE/CVF International Conference on Computer Vision}, 2023, pp. 6166--6176.

\bibitem{neftci2019surrogate}
E.~O. Neftci, H.~Mostafa, and F.~Zenke, ``Surrogate gradient learning in spiking neural networks: Bringing the power of gradient-based optimization to spiking neural networks,'' \emph{IEEE Signal Processing Magazine}, vol.~36, no.~6, pp. 51--63, 2019.

\bibitem{adLIF}
A.~Bittar and P.~N. Garner, ``A surrogate gradient spiking baseline for speech command recognition,'' \emph{Frontiers in Neuroscience}, vol.~16, p. 865897, 2022.

\bibitem{TCLIF}
S.~Zhang, Q.~Yang, C.~Ma, J.~Wu, H.~Li, and K.~C. Tan, ``{TC-LIF}: A two-compartment spiking neuron model for long-term sequential modelling,'' in \emph{Proceedings of the AAAI Conference on Artificial Intelligence}, vol.~38, no.~15, 2024, pp. 16\,838--16\,847.

\bibitem{LMH}
Z.~Hao, X.~Shi, Z.~Huang, T.~Bu, Z.~Yu, and T.~Huang, ``A progressive training framework for spiking neural networks with learnable multi-hierarchical model,'' in \emph{The Twelfth International Conference on Learning Representations}, 2023.

\bibitem{PLIF}
W.~Fang, Z.~Yu, Y.~Chen, T.~Masquelier, T.~Huang, and Y.~Tian, ``Incorporating learnable membrane time constant to enhance learning of spiking neural networks,'' in \emph{Proceedings of the IEEE/CVF International Conference on Computer Vision}, 2021, pp. 2661--2671.

\bibitem{GLIF}
X.~Yao, F.~Li, Z.~Mo, and J.~Cheng, ``{GLIF}: A unified gated leaky integrate-and-fire neuron for spiking neural networks,'' \emph{Advances in Neural Information Processing Systems}, vol.~35, pp. 32\,160--32\,171, 2022.

\bibitem{PMSN}
X.~Chen, J.~Wu, C.~Ma, Y.~Yan, Y.~Wu, and K.~C. Tan, ``{PMSN:} {A} parallel multi-compartment spiking neuron for multi-scale temporal processing,'' \emph{CoRR}, vol. abs/2408.14917, 2024.

\bibitem{ltc}
B.~Yin, F.~Corradi, and S.~M. Boht{\'e}, ``Accurate online training of dynamical spiking neural networks through forward propagation through time,'' \emph{Nature Machine Intelligence}, vol.~5, no.~5, pp. 518--527, 2023.

\bibitem{dhsnn}
H.~Zheng, Z.~Zheng, R.~Hu, B.~Xiao, Y.~Wu, F.~Yu, X.~Liu, G.~Li, and L.~Deng, ``Temporal dendritic heterogeneity incorporated with spiking neural networks for learning multi-timescale dynamics,'' \emph{Nature Communications}, vol.~15, no.~1, p. 277, 2024.

\bibitem{lif}
A.~N. Burkitt, ``A review of the integrate-and-fire neuron model: I. homogeneous synaptic input,'' \emph{Biological Cybernetics}, vol.~95, pp. 1--19, 2006.

\bibitem{duan2022temporal}
C.~Duan, J.~Ding, S.~Chen, Z.~Yu, and T.~Huang, ``Temporal effective batch normalization in spiking neural networks,'' \emph{Advances in Neural Information Processing Systems}, vol.~35, pp. 34\,377--34\,390, 2022.

\bibitem{ASGL}
Z.~Wang, R.~Jiang, S.~Lian, R.~Yan, and H.~Tang, ``Adaptive smoothing gradient learning for spiking neural networks,'' in \emph{International Conference on Machine Learning}.\hskip 1em plus 0.5em minus 0.4em\relax PMLR, 2023, pp. 35\,798--35\,816.

\bibitem{marcus1993building}
M.~Marcus, B.~Santorini, and M.~A. Marcinkiewicz, ``Building a large annotated corpus of english: The {Penn Treebank},'' \emph{Computational Linguistics}, vol.~19, no.~2, pp. 313--330, 1993.

\bibitem{bellec2018long}
G.~Bellec, D.~Salaj, A.~Subramoney, R.~Legenstein, and W.~Maass, ``Long short-term memory and learning-to-learn in networks of spiking neurons,'' \emph{Advances in Neural Information Processing Systems}, vol.~31, 2018.

\bibitem{bellec2020solution}
G.~Bellec, F.~Scherr, A.~Subramoney, E.~Hajek, D.~Salaj, R.~Legenstein, and W.~Maass, ``A solution to the learning dilemma for recurrent networks of spiking neurons,'' \emph{Nature Communications}, vol.~11, no.~1, p. 3625, 2020.

\bibitem{ALIF}
B.~Yin, F.~Corradi, and S.~M. Boht{\'e}, ``Accurate and efficient time-domain classification with adaptive spiking recurrent neural networks,'' \emph{Nature Machine Intelligence}, vol.~3, no.~10, pp. 905--913, 2021.

\bibitem{8891809}
E.~O. Neftci, H.~Mostafa, and F.~Zenke, ``Surrogate gradient learning in spiking neural networks: Bringing the power of gradient-based optimization to spiking neural networks,'' \emph{IEEE Signal Processing Magazine}, vol.~36, no.~6, pp. 51--63, 2019.

\bibitem{bengio2013estimating}
Y.~Bengio, N.~L{\'{e}}onard, and A.~C. Courville, ``Estimating or propagating gradients through stochastic neurons for conditional computation,'' \emph{CoRR}, vol. abs/1308.3432, 2013.

\bibitem{zheng2021going}
H.~Zheng, Y.~Wu, L.~Deng, Y.~Hu, and G.~Li, ``Going deeper with directly-trained larger spiking neural networks,'' in \emph{Proceedings of the AAAI Conference on Artificial Intelligence}, vol.~35, no.~12, 2021, pp. 11\,062--11\,070.

\bibitem{ba2016layer}
L.~J. Ba, J.~R. Kiros, and G.~E. Hinton, ``Layer normalization,'' \emph{CoRR}, vol. abs/1607.06450, 2016.

\bibitem{ioffe2015batch}
S.~Ioffe and C.~Szegedy, ``Batch normalization: Accelerating deep network training by reducing internal covariate shift,'' in \emph{Proceedings of the 32nd International Conference on Machine Learning, {ICML}}, ser. {JMLR} Workshop and Conference Proceedings, vol.~37.\hskip 1em plus 0.5em minus 0.4em\relax JMLR.org, 2015, pp. 448--456.

\bibitem{PSN}
W.~Fang, Z.~Yu, Z.~Zhou, D.~Chen, Y.~Chen, Z.~Ma, T.~Masquelier, and Y.~Tian, ``Parallel spiking neurons with high efficiency and ability to learn long-term dependencies,'' in \emph{Advances in Neural Information Processing Systems}, vol.~36, 2023, pp. 53\,674--53\,687.

\bibitem{CLIF}
Y.~Huang, X.~Lin, H.~Ren, H.~Fu, Y.~Zhou, Z.~Liu, B.~Pan, and B.~Cheng, ``{CLIF}: Complementary leaky integrate-and-fire neuron for spiking neural networks,'' in \emph{Proceedings of the 41st International Conference on Machine Learning}, ser. Proceedings of Machine Learning Research, vol. 235.\hskip 1em plus 0.5em minus 0.4em\relax PMLR, 21--27 Jul 2024, pp. 19\,949--19\,972.

\bibitem{CELIF}
X.~Chen, J.~Wu, H.~Tang, Q.~Ren, and K.~C. Tan, ``Unleashing the potential of spiking neural networks for sequential modeling with contextual embedding,'' \emph{CoRR}, vol. abs/2308.15150, 2023.

\bibitem{DBLP:conf/iclr/HammouamriHM24}
I.~Hammouamri, I.~K. Hassani, and T.~Masquelier, ``Learning delays in spiking neural networks using dilated convolutions with learnable spacings,'' in \emph{The Twelfth International Conference on Learning Representations, {ICLR}}.\hskip 1em plus 0.5em minus 0.4em\relax OpenReview.net, 2024.

\bibitem{hao2024towards}
X.~Hao, C.~Ma, Q.~Yang, J.~Wu, and K.~C. Tan, ``Towards ultra-low-power neuromorphic speech enhancement with spiking-fullsubnet,'' \emph{CoRR}, vol. abs/2410.04785, 2024.

\bibitem{yao2024spike}
M.~Yao, J.~Hu, Z.~Zhou, L.~Yuan, Y.~Tian, B.~Xu, and G.~Li, ``Spike-driven transformer,'' \emph{Advances in Neural Information Processing Systems}, vol.~36, 2024.

\bibitem{2023spikingjelly}
W.~Fang, Y.~Chen, J.~Ding, Z.~Yu, T.~Masquelier, D.~Chen, L.~Huang, H.~Zhou, G.~Li, and Y.~Tian, ``{SpikingJelly}: {An} open-source machine learning infrastructure platform for spike-based intelligence,'' \emph{Science Advances}, vol.~9, no.~40, p. eadi1480, 2023.

\bibitem{Yin2021accurate}
B.~Yin, F.~Corradi, and S.~M. Boht{\'e}, ``Accurate and efficient time-domain classification with adaptive spiking recurrent neural networks,'' \emph{Nature Machine Intelligence}, vol.~3, no.~10, pp. 905--913, Oct 2021.

\bibitem{loshchilov2017fixing}
I.~Loshchilov and F.~Hutter, ``Fixing weight decay regularization in adam,'' \emph{CoRR}, vol. abs/1711.05101, 2017.

\bibitem{merityRegOpt}
S.~Merity, N.~S. Keskar, and R.~Socher, ``Regularizing and optimizing {LSTM} language models,'' in \emph{6th International Conference on Learning Representations, {ICLR}}.\hskip 1em plus 0.5em minus 0.4em\relax OpenReview.net, 2018.

\bibitem{CMOS}
M.~Horowitz, ``1.1 computing's energy problem (and what we can do about it),'' in \emph{2014 IEEE International Solid-State Circuits Conference Digest of Technical Papers (ISSCC)}.\hskip 1em plus 0.5em minus 0.4em\relax IEEE, 2014, pp. 10--14.

\bibitem{snn_temporal_fusion_2024}
Y.~Li, J.~Li, K.~Sun, L.~Leng, and R.~Cheng, ``Towards scalable {GPU}-accelerated {SNN} training via temporal fusion,'' in \emph{Artificial Neural Networks and Machine Learning -- ICANN}.\hskip 1em plus 0.5em minus 0.4em\relax Cham: Springer Nature Switzerland, 2024, pp. 58--73.

\end{thebibliography}

\clearpage
\onecolumn

\appendices


\setcounter{page}{1}

\Large
\begin{center}
    {\bf \LARGE Appendix}\\
    \vspace{0.2cm}
    {\large ``Spiking Neural Networks for Temporal Processing: Status Quo and Future Prospects''}\\
	\vspace{0.2cm}
    {\normalsize Chenxiang~Ma{*}, Xinyi Chen{*}, Yanchen Li{*}, Qu Yang, Yujie~Wu, Guoqi~Li,~\IEEEmembership{Member,~IEEE}, Gang~Pan,~\IEEEmembership{Senior Member,~IEEE},  Huajin~Tang,~\IEEEmembership{Senior Member,~IEEE}, Kay~Chen~Tan,~\IEEEmembership{Fellow,~IEEE}, Jibin~Wu,~\IEEEmembership{Member,~IEEE}}\\
\end{center}


\small

This appendix provides the implementation details, the organization of our source code, and supplementary visualizations for the analysis of the DvsGesture dataset.

\section{Implementation Details of Section~\ref{sec:eval_bench} ``Evaluation of Neuromorphic Benchmarks Using STP''}
\label{app:imp_det_exsiting_bench}
\subsection{Datasets}
We analyze the effectiveness of ten widely used neuromorphic benchmarks in the evaluation of temporal processing capabilities, including three static image recognition datasets (i.e., MNIST~\cite{lecun1998gradient}, CIFAR10~\cite{krizhevsky2009learning}, and CIFAR100~\cite{krizhevsky2009learning}), three event-based vision datasets (i.e., N-MNIST~\cite{orchard2015converting}, DVS-CIFAR10~\cite{li2017cifar10}, and DvsGesture~\cite{amir2017low}), and four audio classification datasets (i.e., GSC~\cite{warden2018speech}, SHD~\cite{cramer2020heidelberg}, SSC~\cite{cramer2020heidelberg}, and TIMIT~\cite{TIMIT}). Details of these datasets and the data augmentation techniques adopted in our experiments are provided below.

\begin{itemize}
    \item \textbf{MNIST} dataset consists of $60,\!000$ training and $10,\!000$ testing handwritten-digit images in $10$ classes. The size of each image is $28\times28$.
    \item \textbf{CIFAR10} and \textbf{CIFAR100} datasets contain a total of $50,\!000$ training images and $10,\!000$ test images, categorized into $10$ and $100$ classes, respectively. For the training set, we apply standard data augmentation, which includes padding each sample with $4$ pixels on all sides, followed by a $32\times32$ crop and a random horizontal flip. In line with previous studies~\cite{ASGL}, we also utilize the AutoAugment and Cutout techniques for additional data augmentation.
    \item \textbf{N-MNIST} dataset is created by capturing static MNIST images with a DVS camera. Each spike pattern has a spatial dimension of $34\times34\times2$ and lasts for $300$ time steps. No data augmentation methods are used.
    \item \textbf{CIFAR10-DVS} dataset is obtained from the CIFAR-10 dataset by scanning each image with repeated closed-loop movements in front of a DVS camera. It includes $9,\!000$ training samples and $1,\!000$ testing samples, with a spatial resolution of $128\times128$. Like CIFAR-10, CIFAR10-DVS comprises $10$ classes. We utilize the standard preprocessing pipeline from SpikingJelly~\cite{2023spikingjelly} to convert events into frames for further analysis, without applying any data augmentation techniques.
    \item \textbf{DvsGesture} dataset comprises $11$ different hand gestures performed by $29$ different subjects, captured using a DVS camera under three varying illumination conditions. We apply the standard preprocessing pipeline from SpikingJelly to transform events into frames with $20$ time steps, without implementing any data augmentation techniques.
    \item \textbf{GSCv2} dataset consists of $105,\!829$ one-second audio clips featuring $35$ different spoken commands, which are recorded by various speakers in diverse environments. Following existing studies~\cite{TCLIF,ALIF}, the original classes are restructured into $12$ categories, including ten words: ``yes'', ``no'', ``up'', ``down'', ``left'', ``right'', ``on'', ``off'', ``stop'', and ``go'', along with a special ``unknown'' class that encompasses the remaining commands, plus an extra ``silence`` class extracted from background noise. The preprocessing methods align with those used in prior studies~\cite{TCLIF,ALIF}.
    \item \textbf{SHD} dataset is a keyword spotting task for spike-based audio classification, featuring spoken digits from $0$ to $9$ in both English and German, which are categorized into $20$ classes. The dataset comprises recordings from twelve distinct speakers, with two exclusively in the test set. Based on the method from previous work~\cite{TCLIF}, each original waveform has been transformed into spike trains across $700$ input channels. The training set includes $8,\!332$ samples, while the test set contains $2,\!088$ samples, with no separate validation set. 
    \item \textbf{SSC} dataset is created from the GSCv2 dataset, utilizing a biologically inspired encoding method to represent data in a spike-based format. It encompasses 35 classes contributed by a diverse array of speakers. In accordance with the method outlined by Zhang~\emph{et al.}~\cite{TCLIF}, the original waveforms have been converted into spike trains across $700$ input channels. The dataset includes $75,\!466$ samples in the training set and $20,\!382$ samples in the test set.
    \item \textbf{TIMIT} dataset is a corpus of read speech that includes time-aligned orthographic, phonetic, and word transcriptions, as well as a single-channel, 16-bit, \SI{16}{\kilo\hertz} speech waveform file for each utterance. It features broadband recordings from 630 speakers --- approximately 70\% male and 30\% female --- representing 8 major dialects of American English, with each speaker reading 10 phonetically rich sentences. Following previous work~\cite{Yin2021accurate}, the raw audio data is preprocessed into Mel Frequency Cepstral Coefficients and converted into frames with 39 input channels. Each frame is then classified into one of 61 phoneme classes for prediction.
\end{itemize}

\subsection{Training Configurations}
Our training configurations are detailed in Table \ref{tab:train_details_benchmarks}. For clarity, the following notations are employed in denoting network architectures: \texttt{C} represents a convolutional layer, \texttt{AP} stands for average pooling, and \texttt{FC} denotes a fully connected layer. Furthermore, \texttt{C3} indicates a convolutional of kernel size $3\times3$. The numerical value preceding \texttt{C} and \texttt{FC} indicates the number of output channels.

\section{Implementation Details of Section~\ref{sec:benchmark} ``Temporal Processing Benchmark Suite''}
\label{app:imp_det_seq_bench}

The training configurations are outlined in Table~\ref{tab:train_details_benchmarks}. Specifically, for the PS-MNIST and Binary Adding tasks, we utilize the AdamW optimizer \cite{loshchilov2017fixing} in conjunction with a step scheduler that reduces the learning rate by a factor of 0.8 every 10 epochs. In contrast, the PTB task employs the stochastic gradient descent (SGD) optimizer without a scheduler. This configuration has been validated as effective for training the SNN methods used in this study. Fully connected network architectures are implemented for all SNNs in the experiments. Consistent with standard practices, the hidden dimensions are set to \texttt{64FC-256FC-256FC} for the PS-MNIST task \cite{ALIF, TCLIF} and \texttt{400FC-1100FC} for the PTB task \cite{merityRegOpt}. For our proposed Binary Adding task, the hidden dimensions are configured as \texttt{128FC-128FC}.


\begin{table}[!t]
\caption{Training configurations and hyper-parameters for the evaluation of neuromorphic benchmarks}
\label{tab:train_details_benchmarks}
\resizebox{\textwidth}{!}{%
\begin{tabular}{lccccccccc}
\hline
  \textbf{Dataset} &
  \textbf{Epochs} &
  \textbf{Optimizer} &
  \begin{tabular}[c]{@{}c@{}}\textbf{Learning} \\ \textbf{Rate}\end{tabular} &
  \begin{tabular}[c]{@{}c@{}}\textbf{Learning Rate} \\ \textbf{Schedule}\end{tabular} &
  \begin{tabular}[c]{@{}c@{}}\textbf{Batch} \\ \textbf{Size}\end{tabular} &
   \begin{tabular}[c]{@{}c@{}}\textbf{Time} \\ \textbf{Step} (\(T\))\end{tabular} &
  \begin{tabular}[c]{@{}c@{}}\textbf{Neuronal}\\ \textbf{Decay}\end{tabular} &
  \textbf{Threshold} &
  \begin{tabular}[c]{@{}c@{}}\textbf{Network} \\ \textbf{Architecture}\end{tabular}\\ \hline
MNIST            & 100 & AdamW   & 0.0001  & Cosine Annealing & 256 & 10 & 0.5 & 0.3 &  \begin{tabular}[c]{@{}c@{}}\texttt{32C3-AP2-32C3-} \\ \texttt{AP2-128FC-10FC} \end{tabular}  \\
CIFAR10          & 200 & SGD   & 0.1 & Cosine Annealing & 128 & 4 & 0.3 & 1.0 & ResNet18  \\
CIFAR100         & 200 & SGD   & 0.1 & Cosine Annealing & 128 & 4 & 0.3 & 1.0 & ResNet18  \\
N-MNIST          & 100 & AdamW & 0.0001 & Cosine Annealing & 16 & 300  & 0.65 & 0.3 & \begin{tabular}[c]{@{}c@{}}\texttt{64C7-AP2-128C7-} \\ \texttt{128C7-AP2-10FC} \end{tabular}   \\
DVS-CIFAR10      &  200  & AdamW & 0.001 & Cosine Annealing & 64 &  10   & 0.3 & 1.0 & VGG11  \\
DvsGesture        & 200 & AdamW & 0.001 & Cosine Annealing & 32 & 20 & 0.3 & 1.0 & VGG11 \\
GSC       & 200 & AdamW & 0.01 & Cosine Annealing & 128 & 101  & 0.8 & 0.5 &  \texttt{(512FC)*6-10FC}  \\ 
SHD       & 200 & AdamW & 0.0005 & -        & 256 & 250  & 0.9 & 0.7 & \texttt{(128FC)*5-20FC}  \\
SSC       & 200 & AdamW & 0.0005 & -        & 256 & 250  & 0.9 & 0.9 & \texttt{(128FC)*5-35FC}  \\ 
TIMIT     & 200 & AdamW & 0.0005 & -        & 64  & 100  & 0.7 & 0.7 & \texttt{(1024FC)*5-61FC} \\ 
\hline
PTB       & 100 & SGD  & 3     & -          & 20  & 70 & 0.5 & 0.5 & \texttt{400FC-1100FC-400FC} \\
PS-MNIST & 100 & AdamW & 0.0005 & StepLR    & 256 & 784 & 1.0 & 0.5 & \texttt{64FC-256FC-256FC-10FC} \\
Binary Adding & 50 & AdamW & 0.0005 & StepLR & 250 & 100 & 0.98 & 0.5 & \texttt{128FC-128FC-10FC} \\ 
\hline
\end{tabular}%
}
\end{table}

\section{Details of Energy Cost Computation for Non-spiking and Spiking Neural Architectures}
\label{sec:energy_·cost}
This section outlines the computation of empirical energy cost metrics presented in Table \ref{tab:energy}. 
We first establish theoretical energy cost formulas for each hidden layer of compared models, as presented in Table \ref{tab:energy_for}. Following standard computational cost evaluation approaches for ANNs and SNNs, we count their number of 32-bit floating-point Multiply-Accumulate (MAC) and Accumulate (AC) operations per inference based on their spatiotemporal dynamics. These numbers of operations are then used to estimate the energy consumption of models on neuromorphic hardware, where $E_\text{MAC}$ and $E_\text{AC}$ denote the energy required per MAC and AC operation, respectively. In these formulas, $m$ and $n$ denote the input sizes, and hidden sizes for each layer, respectively. The variables $f_\text{in}$ and $f_\text{out}$ represent the spike frequencies of input and output spike sequences. For the SpikingTCN, $k$ denotes the convolution kernel size, $f_\text{conv2}$ specifies the spike frequency of input spikes in the second convolution layer. In the Spike-Driven Transformer, $h$ is the hidden dimension of feedforward modules, $f_{Q}$, $f_{K}$, $f_{V}$, $f_\text{attn}$, $f_\text{fc1}$, and $f_\text{fc2}$ denote the spike frequency of neurons in the self-attention and feedforward modules. These spike frequency data is recorded during inference across three tasks and used to compute the number of AC operations.

Subsequently, based on data derived from a \SI{45}{\nano\meter} CMOS process~\cite{CMOS}, the energy per operation is measured to $E_\text{AC} = \SI{0.9}{\pico\joule}$ for AC operations and $E_\text{MAC} = \SI{4.6}{\pico\joule}$ for MAC operations. All the above data are ultimately substituted into the theoretical energy cost formulas, computing the empirical energy cost for each layer. The total energy cost is then obtained by summing the energy consumption across all hidden layers, as demonstrated in the main text.

\begin{table}[h!]
\centering
\caption{Theoretical energy costs of different neural architectures}
\setlength{\tabcolsep}{26pt}
\renewcommand{\arraystretch}{1.25}
\begin{threeparttable}
    \begin{tabular}{l | c}
    \hline
    \textbf{Architecture} & \textbf{Theoretical Energy Cost} \\ \hline
    TCN  &\thead{$\left( kmn + kn^2 \right) \cdot E_\text{MAC}$}  \\ 
    SpikingTCN  &\thead{$\left( kmn \cdot f_\text{in} + kn^2 \cdot f_\text{conv2} \right) \cdot E_\text{AC}$}  \\ 
    LSTM  & \thead{$\left( 4mn + 4n^2 + 19n \right) \cdot E_\text{MAC}$} \\ 
    GSU  & \thead{$\left(2mn \cdot f_\text{in} + 2n^2 \cdot f_\text{out} \right) \cdot E_\text{AC} + 5n \cdot E_\text{MAC} $} \\ 
    Transformer & \thead{$ \left( 4n^2 + 2nT +2nh \right) \cdot E_\text{MAC}$}\\
    Spike-Driven Transformer ($T_{\text{in}}=4$) & \thead{$\left(\left( 12f_\text{in} + 4f_\text{attn} \right) \cdot n^2 +\left( 4f_{Q} \cdot f_{K} + 4f_{V} \right) \cdot nT + \left(4f_\text{fc1} + 4f_\text{fc2} \right) \cdot nh \right) \cdot E_\text{AC} + \left( 24n+4h\right) \cdot E_\text{MAC}$} \\ 
    Spike-Driven Transformer ($T_{\text{in}}=1$)  & \thead{$\left(\left( 3f_\text{in} + f_\text{attn} \right) \cdot n^2 +\left( f_{Q} \cdot f_{K} + f_{V} \right) \cdot nT + \left(f_\text{fc1} + f_\text{fc2} \right) \cdot nh \right) \cdot E_\text{AC}$} \\ 
    \hline
    \end{tabular}
\label{tab:energy_for}
\end{threeparttable}
\end{table}

\section{ Organization of the Benchmarking Library}

In this section, we provide a detailed description of the structure of our open-sourced benchmarking library. Initially, we present an overview of the library, highlighting its main components such as the framework and experimental configurations. We then examine the structure of the model design within the framework, elucidating the role and functionality of each component. This systematic exposition is designed to facilitate the rigorous evaluation and further development of SNN models and datasets in future studies.

\subsection{Overview of the Benchmarking Library}
This subsection presents the architecture of our benchmarking library, which comprises two key components: \texttt{framework} and \texttt{experiments}. Together, these components establish a comprehensive structure that supports both model development and task design while ensuring a well-defined experimental setup. They collectively provide essential interfaces that enable users to interact with the benchmark and contribute new features.

The \texttt{framework} serves as the foundation for model design, offering functionalities for defining SNN models and selecting appropriate training strategies. This component plays a crucial role in deploying models for specific tasks and incorporates Compute Unified Device Architecture (CUDA)-based acceleration kernels along with preprocessing capabilities tailored for task-specific datasets.

The \texttt{experiments} component provides a detailed set of benchmark experiments, serving as practical references for users. Each experiment is organized as an independent module, consisting of a code file and a configuration file. The code file specifies the experimental workflow, while the configuration file simplifies the process by enabling users to load and manage experiment-specific configurations, including hyperparameter settings efficiently.


\subsection{Structure of \texttt{framework}}

The \texttt{framework} component of the benchmarking library is designed to assist users in defining, training, and deploying SNN models through structured interfaces. It consists of three key modules: \texttt{kernel}, \texttt{network}, and \texttt{utils}. A detailed breakdown of these components and their specific functionalities is presented in Table~\ref{tab:framework}. The \texttt{kernel} module facilitates CUDA-based acceleration to enhance computational efficiency, incorporating an optional kernel based on the temporal fusion method~\cite{snn_temporal_fusion_2024}, which is specifically designed to accelerate processing within spiking neuron layers during long temporal sequences in both training and inference.

Within the \texttt{network} module, three primary components are provided: \texttt{neuron}, \texttt{structure}, and \texttt{trainer}.
The \texttt{neuron} component enables users to select particular spiking neurons.
The \texttt{structure} component offers a collection of predefined network architectures.
Once the model structure is established, the \texttt{trainer} component provides various training strategies to optimize model performance.

The \texttt{utils} module offers essential functionalities for both training and deployment.
It includes the \texttt{tools} component, which provides optional utilities to facilitate model training, and the \texttt{dataset} component, which supports various neuromorphic datasets.



\begin{table}
\caption{Overview of the benchmarking library}
\label{tab:framework}
    \setlength{\tabcolsep}{14.2pt}
    \renewcommand{\arraystretch}{1.25}
    \centering
    \begin{tabular}{l|c|c|c}
    \hline
    \textbf{Module} & \textbf{Component} & \textbf{Instance} & \textbf{Description} \\
    \hline
    \texttt{kernel} & - & \texttt{temporal\_fusion\_kernel}, etc. & \makecell[c]{Ready-to-use accelerated CUDA kernel wrappers \\ optimized for spiking neurons.} \\
    \hline
    \multirow{3}{*}{\texttt{network}} & \texttt{neuron} &  LIF, ALIF, TDBN, TEBN, etc. & \multirow{3}{*}{\makecell[c]{Interfaces for modules defining network layers, architectures, \\ and the corresponding training methods required for SNNs.}} \\
     & \texttt{structure} & GSU, SpikingTCN, etc. & \\
     & \texttt{trainer} & \texttt{SurrogateGradient}, etc. & \\
    \hline
    \multirow{2}{*}{
    \texttt{utils}} & \texttt{dataset} & PTB, PS-MNIST, etc. 
    & \multirow{2}{*}{\makecell[c]{Encapsulation of utility functions and optimized dataset \\ modules for efficient model training and validation.}} \\
    & \texttt{tools} & \texttt{logging}, \texttt{save\_checkpoint}, etc. & \\
    \hline
    \end{tabular}
\end{table}

\begin{figure*}[!htb]
\centering\includegraphics[width=0.97\linewidth]{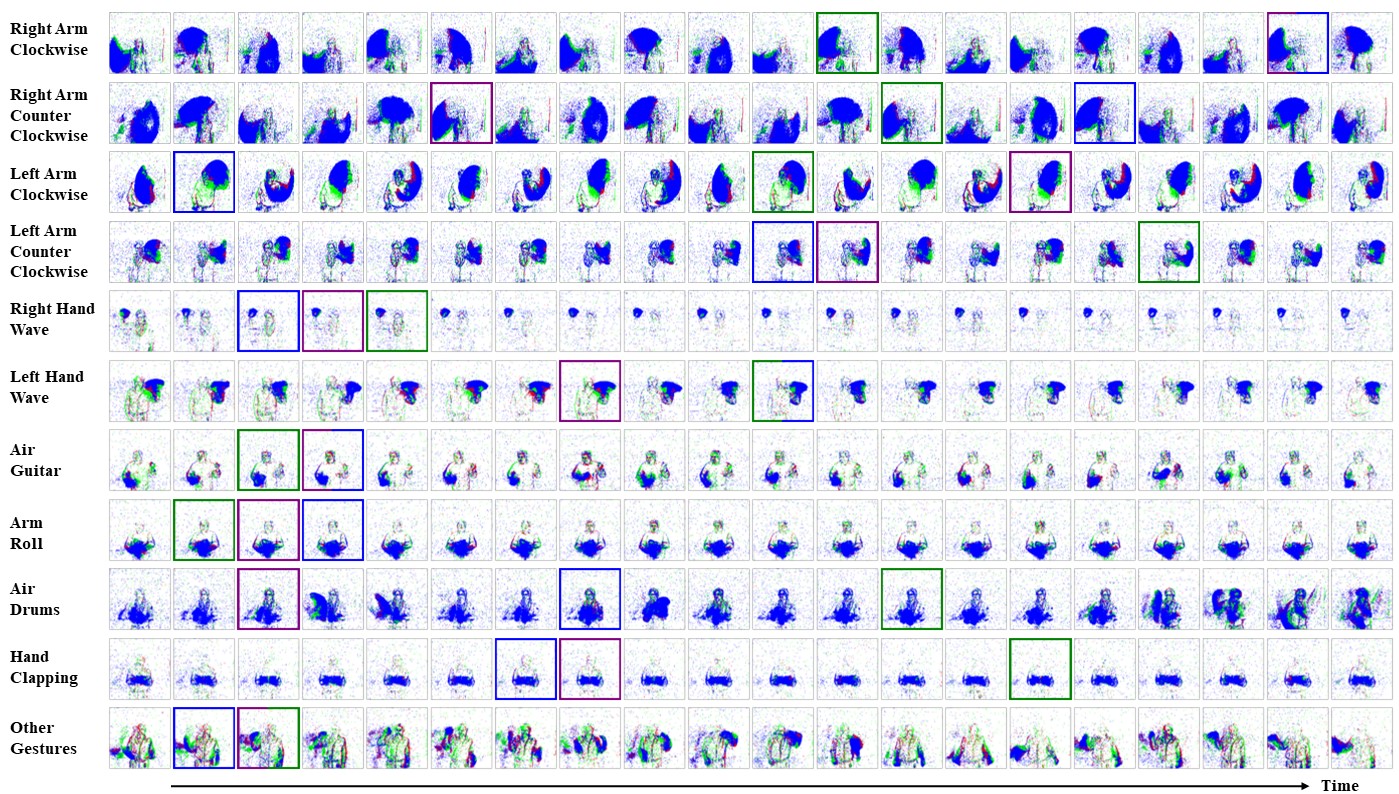}
\caption{Qualitative results of samples from the DvsGesture dataset. In these samples, the most confident frame differs across the three algorithms in STP. Frames selected by STBP are highlighted with purple boxes, SDBP with green boxes, and NoTD with blue boxes.}
\label{Fig:app_dvs_vis_diff_t}
\end{figure*}

\begin{figure*}[!htb]
\centering\includegraphics[width=0.97\linewidth]{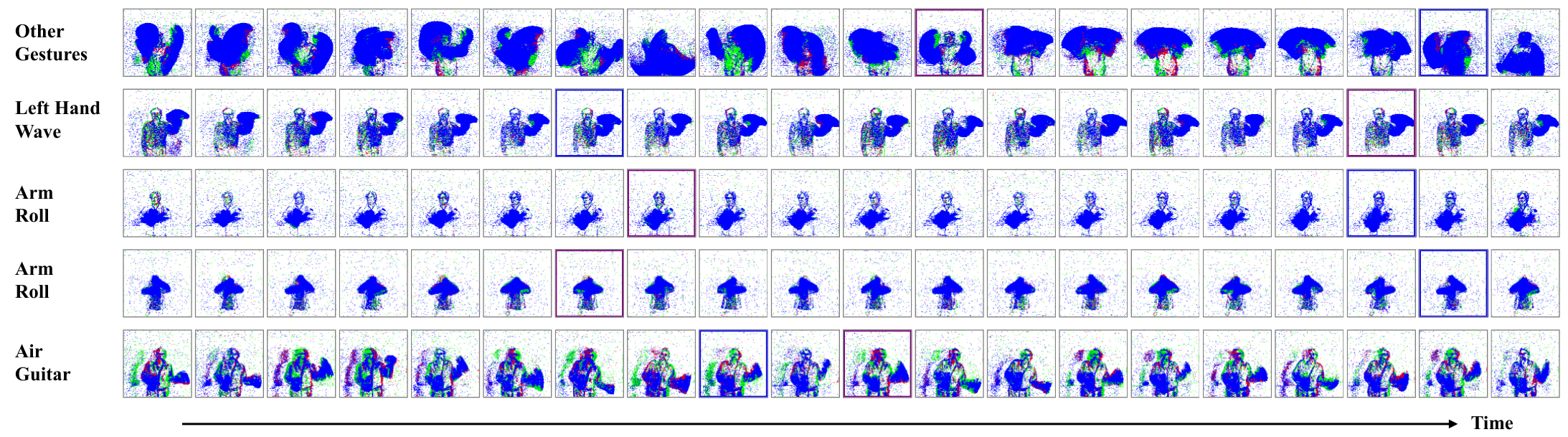}
\caption{Visualization of samples from the DvsGesture dataset that STBP classifies correctly, while NoTD does not. The predictions by NoTD are right arm counter clockwise, left arm clockwise, air drums, hand clapping, and other gestures (from top to bottom).}
\label{Fig:app_dvs_vis_stbp_correct_notd_wrong}
\end{figure*}

\begin{figure*}[!htb]
\centering\includegraphics[width=0.97\linewidth]{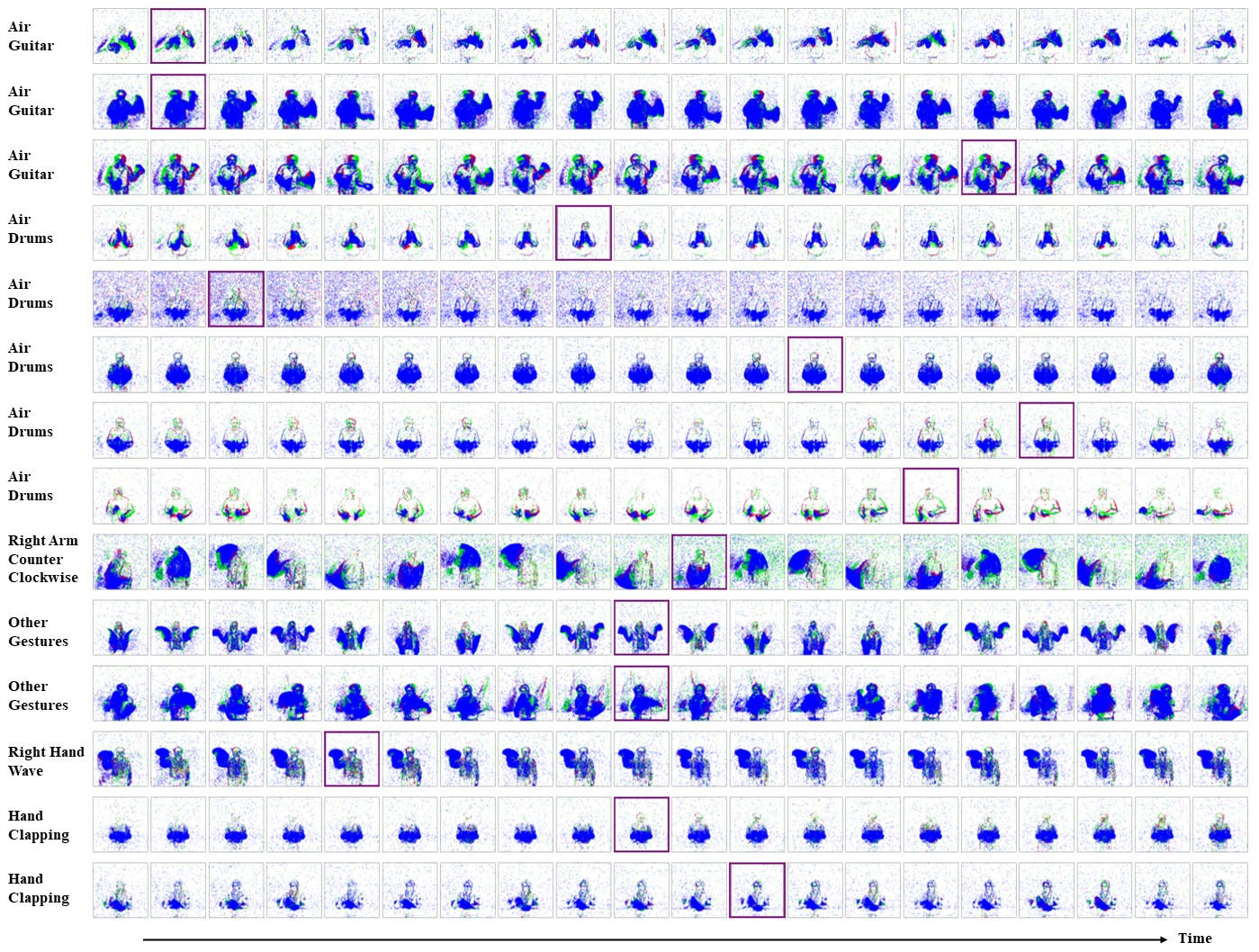}
\caption{Visualization of samples from the DvsGesture dataset that are misclassified by STBP.  The predictions by STBP are other gestures, air drums, other gestures, hand clapping, arm roll, arm roll, hand clapping, other gestures, other gestures, air guitar, air drums, right arm counter clockwise, arm roll, air drums (from top to bottom).
}
\label{Fig:app_dvs_vis_stbpwrong}
\end{figure*}

 




\vfill

\end{document}